\documentclass[conference,12pt,a4paper]{IEEEtran}

\usepackage{flushend}  
\usepackage{ulem}
\usepackage{alphalph}
\usepackage[unicode=true, bookmarks=true,bookmarksnumbered=false,
            bookmarksopen=false,hyperfootnotes=false,breaklinks=true, pdfborder={0 0 1},backref=false,colorlinks=true]{hyperref}
\usepackage{amsmath}
\usepackage{amssymb}
\usepackage{amsthm}

\IEEEoverridecommandlockouts 

\begin{document}
\title{A Chronological Survey of Theoretical Advancements in Generative
Adversarial Networks for Computer Vision}

\author{\IEEEauthorblockN{Hrishikesh Sharma,~\IEEEmembership{Member,~IEEE,}}
\IEEEauthorblockA{\textit{TCS Research} \\
hrishikesh.sharma@tcs.com}
}

\maketitle

\begin{abstract}
Generative Adversarial Networks (GANs) have been workhorse generative
models for last many years, especially in the research field of computer vision.
Accordingly, there have been many significant advancements in the
theory and application of GAN models, which are notoriously hard to train,
but produce good results if trained well. There have been many a surveys on
GANs, organizing the vast GAN literature from various focus and
perspectives. However, none of the surveys brings out the important
chronological aspect: how the multiple challenges of employing GAN models
were solved one-by-one over time, across multiple landmark research works.
This survey intends to bridge that gap and present some of the landmark
research works on the theory and application of GANs, in chronological
order.
\end{abstract}

\section{Introduction}
\label{intro_sec}

\IEEEPARstart{G}{enerative} Adversarial Models have been the mainstay of
deep generative modeling for almost a decade. Even with the advent of
diffusion models whose theory is slowly maturing \cite{latent_diff_pap,
thermo_gen_pap, ddim_pap, sde_dpm_pap, iddpm_pap, guided_diffusion_pap,
grad_gen_pap, ddpm_pap}, GANs remain
the most practically employed generative models in computer vision area,
till date. This is especially true for the conditional variant of the GANs.
Their application area spans from image inpainting, image superresolution,
image-to-image translation, anomaly detection, image
deblurring/dehazing/deraining, image and video synthesis, to name a few.

However, GANs are notoriously hard to train. Being \textbf{implicit}
generative models \cite{dgm_book}, the model attempts to snug fit any
generative probability density function(p.d.f.) of the input dataset. The
do so by trying to game-theoretically achieve the Nash Equilibrium between
a generator and a discriminator network. Especially in case of diverse
datasets having multiple statistical modes in its empirical p.d.f., GANs
face significant challenges in learning all the modes. Most of the time
they learn just the first few modes with higher mass, a problem that is
known as the \textbf{mode collapse} problem. Another problem with GANs,
rather generative models in general, is the ability to generate
high-resolution images. While the problem has eventually been solved, the
solution is based on many a advancements that happened in the interim. To
summarize, use of game theory in generative modeling was an altogether new
approach, which brought with itself many theoretical problems that were
solved one-by-one.

This survey intends to chronicle the most important developments and
advancements that have happened in GAN-based deep generative modeling. Ever
since Ian Goodfellow presented the seminal work a decade ago
\cite{gan_orig_pap}, this avenue of generative modeling has exploded. Originally intended to be trained using
unsupervised learning, some of the most important progresses in generative
modeling using GAN has been via supervised learning (as well as
semi-supervised learning also). The survey ensures to adhere to time order
of the works, so as the exposit the natural progression of research in the
recent direction. The survey additionally brings out important progresses in
certain computer vision applications of GAN theory. Last but not least, the
papers only present the salient aspects of a research work, while the
corresponding code speaks for itself, being the actual model. In places
where code was available, the authors studied the code vis-a-vis the
summary in the paper, and have provided some additional insights about the
corresponding research work, something which no other survey has tried till
date.

\section{Evolution of GAN Research}

In this section, we present the main advancements in GAN modeling over the
last decade, as motivated earlier.

\subsection{Generative Adversarial Nets}
\label{orig_gan_sec}
This is the seminal work which introduced the unsupervised,
\textbf{vanilla} version of GAN framework \cite{gan_orig_pap}. They authors
introduced a  new framework for estimating generative models trained via a
newly-defined adversarial process. Adversarial training includes
simultaneously training of two models. First part of training is that of a
generative model $\mathbb{G}$, that, as always, captures the data
distribution. The other part of training is that of a discriminative model
$\mathbb{D}$ that estimates the probability that a sample came from the
training data rather than $\mathbb{G}$. Such training corresponds to a
minimax two-player game. More specifically, discriminative model
$\mathbb{D}$ learns to determine whether a sample is from the model
distribution or the data distribution.  Competition in this game drives
both teams to improve their methods until the counterfeits are
indistinguishable from the genuine articles. Practically, GANs have limited
capacity and hence the generative model is parametric, mostly a MLP. The
game ends at a saddle point, neither a maxima nor a minima. As a practical
step, authors recommend alternating between k steps of optimizing
$\mathbb{D}$ and one step of optimizing $\mathbb{G}$. As another practical
step, the authors recommend that rather than training $\mathbb{G}$ to
minimize log(1 - $\mathbb{D}(\mathbb{G}(\mathit{z})))$ they can train
$\mathbb{G}$ to maximize log $\mathbb{D}(\mathbb{G}(\mathit{z}))$. The
single biggest theoretical highlight of the paper is the proof that $p_g$
indeed tends towards $p_{data}$, while minimizing the Jenson-Shannon
divergence between the two distributions, under certain conditions listed
in Theorem 1.

\subsection{Conditional Generative Adversarial Nets}
\label{cgan_sec}
In a seminal work \cite{cgan_pap}, the authors introduced the
\textbf{supervised training} way to training GANs. They introduce the usage
of \textit{y} i.e. the observations/labels into the training, other than
the data \text{X} already. \textit{y} y simply is used to condition on to
both the generator and discriminator.  They first describe the drawbacks of
existing training method. According to them, it remains tough to
retarget the unsupervised GAN model to accommodate a high number of output
categories. Another highlighted drawback is that existing supervised models
till date learn deterministic bijective mappings from input to predictions.
However, many important tasks, such as image captioning, in fact entail
probabilistic one-to-many mapping. These problems happen since there is no
control on modes of the output being generated. To overcome these, they
suggest conditioning the generative model on additional supervisory inputs,
which can direct the data generation process. Such conditioning is
generally in form of class labels (most common option), or other options as
given the paper. If one looks carefully, it is their modeling of posterior
predictive distribution that leads to one-to-many mapping.
The authors motivate to perform the conditioning by feeding \textit{y} into
the both the discriminator and generator as additional input layer. This is
something that many later papers have followed in word and spirit. Usage of
\textit{y} in the loss function in equation (2) clearly points out to the
traditional flavor of supervised learning, which unlike unsupervised
learning, is interested in inferring conditional probability distribution
p(\textit{x}/\textit{y}).

\subsubsection{Salient Observations from Code}
We observed that
\begin{enumerate}
    \item Usage of sample\_interval to control the periodicity after which partially trained generator can be used to create samples from latent noise.
    \item Different samples of noise vectors used in generator training
versus generator sampling stages.
    \item Choice of initialization for $\beta$, $\gamma$ and $\epsilon$ in batchNorm2D layers
    \item Concat operation as the way of inserting supervisory signal in both generator and discriminator (see forward() method)
    \item ANN rather than CNN (linear layers). CNN usage started with D``C''GAN.
    \item Usage of leaky ReLU in discriminator ANN. Many future works have
used the same.
    \item Usage of tanh() and sigmoid() for feature projection
    \item BCE Loss for discriminator loss. At all places, where loss is to
be computed, there are two classes of inputs that are to be discriminated.
    \item Usage of ADAM optimizer. ADAM was the choice for GANs, till TTUR work perhaps brought back SGD into fashion.
    \item Usage of detach() in discriminator, to cut off generator autograd
during generator training.
\end{enumerate}

\subsection{Unsupervised Representation Learning With Deep Convolutional Generative Adversarial Networks}
\label{dcgan_sec}
The foundational work in \cite{dcgan_pap} is the first practical and
scalable realization of GANs introduced in subsection \ref{orig_gan_sec},
and a great starting point to understand (unsupervised) GAN training. The
work, called \textbf{DCGAN}, also cross-refers to a set of GAN training
tips' blog by one of the authors (c.f. subsection \ref{tut_sec}). It uses
CNNs instead of MLPs originally promoted by the GAN authors, to realize
both $\mathbb{G}$ and $\mathbb{D}$. This way, the deep convolutional
adversarial pair learns a hierarchy of representations from object parts to
scenes in both the generator and discriminator. They also promote  reusing
parts of the generator and discriminator networks as \textit{feature
extractors} for supervised tasks, post adversarial training. The single
biggest contribution of this paper is that usage of CNN made GAN training
much \textbf{stable}: the original GAN and its few variants were
notoriously unstable during training. Even when training succeeded, the
authors note that the outputs of original GAN were noisy and
incomprehensible. Similarly, they note that the output of LapGAN(c.f.
subsection \ref{lapgan_sec}) had a problem of looking shaky due to
introduction of noise introduced while chaining multiple models. DCGANs
\cite{dcgan_pap} was the first GAN model, which could generate high resolution images, thus improving on LapGAN which are multi-stage generators. The DCGAN
design leverages three particular advancements in the field of CNN at that
time. Another advantage of DCGAN design is that it tends to learn
\textbf{disentangled representations}, something which is highly desired
for network interpretability, a current hot topic. Generally disentangled
learning is covered as part of unsupervised learning.

\subsubsection{Salient Observations from Code}
\begin{enumerate}
    \item Usage of torch.manual\_seed()
    \item Weight initialization is different for conv and BN layers
    \item Usage of ConvTranspose2d for upsampling
    \item LeakyReLU for discriminator
    \item Choice of Adam Optimizer
    \item Deep Supervisory loss calculation for generator
\end{enumerate}

\subsection{Deep Generative Image Models using a Laplacian Pyramid of Adversarial Networks}
\label{lapgan_sec}
An early and seminal work \cite{lapgan_pap} called LapGAN focuses
specifically on generation of \textit{natural} images. According to the
authors, most GAN works till date focussed on generating image patches, and
hence did not scale well to full image level. Their work focused on
generating plausible looking scenes upto size $64\times64$. They employ a
Laplacian image pyramid representation, and use a generative model trained
at each level using conditional adversarial learning, thus building on
cGANs (c.f. subsection \ref{cgan_sec}). As per authors, earlier generative
parametric models which included various kinds of Boltzmann machines and
variational autoencoders, could only generate realistic output on simple
datasets such as MNIST and NORB. Their training complexities limit their
employment for generation of larger and more realistic images. In LapGAN, there is a
set of generative convnet models \{$\mathbb{G_0}$, $\cdots$,
$\mathbb{G_k}$\}. Each model mimics the distribution of coefficients
$h_k$ for natural images at a different level of their Laplacian pyramid.
These $h_k$'s are used in the classical recursive reconstruction method of
images given their Laplacian pyramid representation which is essentially a
sequence of bandpass images, plus a low-frequency residual. Hence the
models at all except the final level are conditional GANs, that input an
additional upsampled form of current image $\tilde{I}_{k+1}$ as a
conditioning input, in addition to the noise input $z_k$. Since each level
of pyramid is independently trained, the overall GAN model finds it much
difficult to memorize the training input samples. The authors also empirically show that data augmentation is a
useful component of GAN training.

\subsubsection{Salient Observations from Code}
\begin{enumerate}
    \item There are multiple python classes of discriminator. They are numbered 0, 1 and 2 for CIFAR10 (2: 8$\times$8, 1: 14$\times$14, 0: 28$\times$28), as per Figure 1 of paper.
    \item ditto for generators
    \item Adam optimizer is the choice again, though paper mentions SGD
    \item Generator subgraph/subnetwork is not detached during backward
pass over discriminator, unlike in DCGAN. Instead, retain\_variables() has
been used!
\end{enumerate}

\subsection{Unsupervised and semi-supervised learning with categorical generative adversarial networks}
\label{catgan_sec}
 A useful initial work that uses GANs for unsupervised learning on a
specific task is CatGAN \cite{catgan_pap}. The work attempts the task of
unsupervised or semi-supervised classification, also mostly known as
clustering/cluster assignment problem. More specifically, they target
discriminative clustering i.e. maximum-margin separation of clusters.
 To be able to classify, the authors make a critical assumption that the
input distribution $p(x)$, learnt during unsupervised learning, contains
information about $p(y|x)$, where $y \in \{1, \ldots, K\}$ denotes the
unknown label. With this assumption, samples drawn from $p(x)$ can be used
learn $p(y|x)$. Such a representation is then expected to help classifiers
trained in few-shot regime, to generalize to those sections of the data
distribution, about which it would have no knowledge about. Those parts can be those unravelled modes of the p.d.f., which are not
covered by the representation behind the few labeled examples. All earlier
works such as DBM, RBM, various autoencoders relate to generative
clustering methods, which are based on reconstruction of training examples.
One drawback of such reconstruction-based learning methods is that, by
definition, they learn representations which preserve entire information
present in the input images. This goal of perfect reconstruction in general
contradicts the goal of learning a discriminative classifier modeling
$p(y|x)$, which only preserve information just enough to predict the class
identity. CatGAN learns a discriminative classifier $\mathbb{D}$ that
maximizes mutual information between the input $x$ and the supervisory
ground truth $y$ (i.e. $p(y|x, D)$) for $K$ unseen classes. To help
$\mathbb{D}$ in its task of finding classes by generalizing well to unseen
data, an auxiliary target of robustness of the classifier to adversarially
generated examples is added. One limitation of this work is that it always assumes a
uniform prior over classes.

\subsection{Generative Visual Manipulation on the Natural Image Manifold}
\label{vis_manip_sec}

As another application of GAN, a well-cited work \cite{vis_manip_pap} deals
with the task of realistic image manipulation and synthesis, something
which was traditionally done using techniques of computer graphics. As per
authors,  even a simple image manipulation in Photoshop presents
enough difficulties. One reason is lack of safe-editing limits in such
techniques: any slightly imperfect edit leads to a completely unrealistic
output immediately. Recent advancements in GAN and thereby generative
modeling has brought about safe and pleasing synthetic image outputs, by
sampling random codes drawn from the natural image manifold and decoding
them. However, two reasons prevent their advantages from being used in
practical applications. One, the generated images are pleasing but not quite
photo-realistic or high-resolution. Two, the latent space is typically
sampled at random, in these generative models, before being decoded into a
synthetic image. Hence these models are not able to provide user-level
control over creation and manipulation of input images. In this paper, the authors use a GAN to learn the manifold of
natural images. They formally define $\tilde{M} = \{G(z)| z \in Z\}$ and
use it as an approximation to the ideal manifold M (i.e $\tilde{M}\approx
M$). However, they do not employ $\tilde{M}$ for image generation. Instead,
they use it to \textit{safe-limit} the output of various image edit
operations. This ensures that the edited output always lies on the learned manifold.
Many editing operations are formulated this natural and data-driven way, specifically
color and shape manipulations. The model then \textit{automatically}
adjusts the output to be as realistic as possible. For a good
initialization of $z$, the training starts from \textit{multiple random
initializations}, and the solution with minimal edit cost is retained.

\subsection{Coupled Generative Adversarial Nets}
\label{coupled_gan_sec}
This work \cite{coupled_gan_pap}, like many other concurrent works, has
been superseded by the seminal work of CycleGAN (c.f. subsection
\ref{cyclegan_sec}). The work picks up the task of unsupervised image
translation. It relies on \textit{weak} supervision: instead of using very
many image pairs for training, it just uses a pair of image domains (two
sets of related images) for supervision of the task. A probability density
function using image samples from two paired domains is essentially a \textbf{joint
distribution}, wherein scenes that are jointly imaged in different domains
are assigned a scalar density value. Examples include images of the same
scene captured from different modalities(depth and RGB images). The joint p.d.f. can also be defined for images of the same object with different attributes
(e.g. angry and worried faces). Once a joint distribution is
learned, the authors use it to synthesize novel pairs of images. They do so
by enforcing the layers that decode high-level semantics in two GANs, to share the
weights in Siamese fashion. This forces the GANs to synthesize the high-level semantics in the same
fashion. The layers that decode low-level details independently map the
output in individual domains to confuse the respective (real-vs-fake) discriminative models.

\subsection{Disentangling factors of variation in deep representations using adversarial training}
\label{disentangle_sec}
Often times, the purpose for which a dataset is collected is to further
progress in solving a certain supervised learning task. Such learning task
is fully label-driven. As a goal, such learning task produces 
representation that remains invariant to the factors of variation that do not
impact the particular supervised learning task.
Ideally, one would like to be able to
learn representations in which the uninformative factors of variation are
separated from the informative ones, instead of being discarded. In a
particular work \cite{disentangle_pap}, a deep conditional generative model
is introduced that learns to
separate the factors of variation leading to formation of the
discriminative labels, from the other unimportant factors of variability.
Such representation helps in certain synthesis tasks, which entail
the use of generative models in a particular way. This includes generating
novel images wherein only certain factors of variations are varied, and
rest are kept fixed, e.g. in apparel recommendation.

\subsection{Improved Techniques for Training GANs}
\label{improved_sec}
The authors of \cite{improved_gan_pap} present a series of new
architectural patterns and training guidelines that apply to GANs. They build it on top of DCGAN innovations (c.f. subsection
\ref{dcgan_sec}). \textbf{Feature matching} tries to overcome the instability of
GANs by employing a new loss function for the generator that stops it from
overtraining on the current discriminator. Instead of directly maximizing
the margin of the discriminator, the new objective forces the generator
to generate fake images that match at the statistics level, the statistics of the real
images.  Specifically, the generator is trained to match the statistically expected value
of the features, at an intermediate layer of the discriminator network.
Minibatch discrimination, one-sided label smoothing etc. are some more
innovations that are motivated and described. This work is foundational and
has been used in multiple subsequent applications of GAN. The authors also
introduce a measure of image generation quality called \textbf{Inception
Score}, something that was widely used and is still a popular metric. Some
limitation of these techniques are quoted in subsection \ref{tripgan_sec}.

\subsubsection{Salient Observations from Code}
\begin{enumerate}
    \item Usage of LinearWeightNorm layer in discriminator and generator
    \item AWGN added to each layer of discriminator
    \item Original code is in Theano. Imagenet-specific experiment is in TF. Pytorch version seems to be highly watered down version, with not all 5 mentioned tricks visible.
\end{enumerate}

\subsection{GAN Tutorial}
\label{tut_sec}

There exists a writeup \cite{gan_tut} corresponding a very well-received
tutorial presented at NIPS'16, by the original authors of GAN. The bring
out certain salient aspects of GAN missed out in the original paper. GAN
perform generative modeling by generating samples from the data
distribution, rather than estimating the distribution itself. They provide
several contemporary reasons about why one must still study generative
modeling. Similarly, they also quote many important and contemporary
applications, including SISR, art synthesis, image translation and domain
transfer. One very important ML principle brought out is that one can also
think of maximum likelihood estimation of model parameters as
\textbf{minimization of KL divergence} between the real data distribution
and distribution of fakes as parameterized by the generative model. GANs
belong to the implicit, direct density model category of generative
modeling, to avoid pitfalls and disadvantages of methods belonging to other
categories. Hence design of GAN entails very few requirements and
assumptions. One constraint this category leads to is that the generator
cannot learn to generate \textbf{discrete output}. For gradient descent in
both $\mathbb{G}$ and $\mathbb{D}$, Adam optimizer is mostly a good choice.
Authors also say that only one step of SGD for $\mathbb{G}$ per one step of
SGD for $\mathbb{D}$ is sufficient. By training the discriminator, the
ratio \[\frac{p_{data}(x)}{p_{model}(x)}\] is estimated at every input
point $x$. Estimating this ratio can be done by employing a wide variety of
divergences as losses. This ratio is susceptible to the problems of
overfitting and underfitting, present in supervised learning. These
problems can be handled in usual way by doing proper optimization and using
high amount of training data. The
authors give detailed reason why the zero-sum game has to be practically
converted into heuristic, non-saturating game. The authors debate the reason for \textbf{mode collapse} problem and argue that
GANs, at times, do not generate a majority of modes of the data
distribution due to problem in their training. This problem occurs when the
generator learns to synthesize several vastly different input random noise samples
into the same decoded output. Hence the applications of GANs till date are
mostly limited to scenarios where it is acceptable for the model to output
a small number of distinct synthetic images, e.g. conditional tasks
including translation and adaptation. One common strategy is to make the
discriminator have more capacity: it is usually deeper and can have more
neurons per layer than the generator.
Authors cross-refer to tips and tricks by DCGAN authors, namely
\begin{enumerate}
    \item train with labels i.e. supervision
    \item Use one-sided label smoothing
    \item perform virtual batch normalization
\end{enumerate}
Authors additionally refer to tips for improved training
\cite{improved_gan_pap} (c.f. subsection \ref{improved_sec}). At the end,
the authors do summarize some of the recent works without criticizing
them.

\subsection{Conditional image generation with PixelCNN decoders}
\label{cgan_pixel_sec}
Many of the practical applications of generative image modeling entail
conditioning of the model on some form of prior knowledge about the input
images, such as class labels. A popular work \cite{pixelcnn_pap}
deals with conditional image generation with a new image density model
based on the PixelCNN \cite{pixelcnn_pap}. The model can be conditioned
using any additional information, including labels or captions,
or even latent embeddings created by various backbones. When conditioned on
class labels, it was able to generate diverse, realistic scenes, e.g.
distinct animals, objects, landscapes and structures present in ImageNet
dataset. When conditioned on a representation produced by a backbone CNN, given a
single object-centric image e.g. that of an unseen face, the model
generated diverse new images of the same person with different face poses,
different ambient lighting and facial expressions.

\subsection{InfoGAN: Interpretable Representation Learning by Information Maximizing Generative Adversarial Nets}
\label{infogan_sec}
The motivation for their investigation is an \textbf{important} summary and
is given below. They define Unsupervised Learning as the problem of
extracting reusable representations from vast amount of unlabeled data
that is available. Such learnt representation should ideally expose
important semantic features as its easily decodable factors. It is also
likely to be useful for many downstream tasks such as classification,
localization, and policy learning in reinforcement learning. Though
this problem is obviously ill-posed since we do not know the relevant
downstream tasks beforehand, during training phase, such decoded factors
within the learnt representation are expected to correspond to a set of
independent manifold dimensions for various attributes such as facial expression, person's identity,
presence of eyeglasses( boolean attribute), hairstyle, eye color etc. Given
such disentangled representation, humans easily perform various tasks that
require knowledge of the salient attributes, such as face and object recognition.
In an inverse sense, for its output representation to be useful, an
unsupervised learning-based model  must be able to guess the plausible
downstream tasks, and create compatible disentangled representations accordingly.

The authors rightly observe that much of unsupervised learning research is driven by
generative modelling. The conjecture is that the ability to synthesize, or
“create” the observed data and similar samples, requires some form of
understanding of the data domain, and that a good generative model hence
automatically learns a disentangled representation.

Improving upon the work in subsection \ref{disentangle_sec},
\cite{infogan_pap} provides an information-theoretic extension to the
Generative Adversarial Network that is able to learn disentangled
representations in a completely unsupervised manner. The author point to
the limitations of earlier GANs in disentangling of representations: they
can only disentangle few of the discrete latent factors, and the
computational complexity is exponentially proportional to the number of factors to
be disentangled. The authors of \cite{infogan_pap} present a simple
variation of the vanilla GAN's objective function, that encourages it to
learn meaningful and disentangled representations. The GAN maximizes the mutual
information between a fixed small subset of the GAN’s input noise variables and the
outputs, which is quite easy to implement. For
example, InfoGAN successfully disentangles writing styles from digit shapes
on the MNIST dataset, pose from lighting of 3D rendered images, and
background digits from the central digit on the SVHN dataset. It also
discovers visual concepts that include hair styles, presence/absence of
eyeglasses, and emotions on the CelebA face dataset. 

\subsection{Adversarially Learned Inference}
\label{ali_sec}
A first work which tried to extend the generator towards an
autoencoder-style architecture for direct supervision is ALI
\cite{ali_pap}. The motivation for generator+inference network can be found
in \cite{gan_overview_pap}. ALI was superseded by ALICE (refer subsection
\ref{alice_sec}), which in turn was bettered by the seminal cycleGAN work.
The method jointly learns a generative network and an encoding/inference
network, while being guided by an adversarial loss. Variational autoencoders, another
contemporary generative model already learn an approximate inference
submodel, which can be reused for various auxiliary tasks. However, the
training of the submodel using MLE runs into
a well-known drawback of blurring. There have been some efforts on bridging
gap between VAE and GAN. ALI fits into that role. It combines the learning of
both an inference submodel (or encoder) and a deep generative submodel
(or decoder) into a single GAN-like model. An additional discriminator is
then trained to discriminate joint samples of the input data and the corresponding
latent variable from the encoder (i.e. approximate posterior), from joint
samples of the synthetic output data and the input noise samples. The
encoder and the decoder jointly train and learn to fool the discriminator, as
is necessary for adversarial learning. Due to joint learning, the
discriminator not only distinguishes synthetic output samples from real
data samples, but it also distinguishes between two joint distributions over
the input data sample space and the latent space. It improves upon InfoGAN (refer
subsection \ref{infogan_sec}) since it only enforces sampling from the
inference submodel, and can hence represent arbitrarily complex
posterior distributions. ALI is strikingly similar to BiGAN work (refer
subsection \ref{bigan_sec}). The main difference is that BiGAN contains a deterministic $q(z|x)$ network, whereas ALI
uses a stochastic network for the same. In authors' experience, this does not make a big
difference practically.

\subsection{Adversarial Feature Learning}
\label{bigan_sec}
Another popular work strikingly similar to ALI, BiGAN \cite{bigan_pap},
tries to answer the question: can GANs be used for unsupervised learning of
reusable feature representations arising from arbitrary input data
distributions? An immediate problem with doing so is that while the
generator maps random latent samples to generated output, there is no
inverse mapping of input data to latent representation in the GANs till
that date. To do so, the authors extend the BiGAN discriminator
$\mathbb{D}$ so that it not only discriminates not only in pixel/data space
($x$ versus $\mathbb{G}(z)$), but jointly in pixel/data and latent space
(pairs ($x$, $\mathbb{E}(x)$) versus ($\mathbb{G}(z)$,
$z$)). The latent part here is either the generator network's input, $z$,
or the encoder network's output, $\mathbb{E}(x)$. The authors argue both
intuitively and formally prove that the encoder and generator network must
stably learn and invert each other's outputs, before they can jointly learn
to fool $\mathbb{D}$, the BiGAN discriminator.

\subsection{Energy-based Generative Adversarial Networks}
\label{ebgan_sec}
Summary-wise, the energy-based model proposed in \cite{ebgan_pap} tries to
approximate a function that maps each point of an input data space to a single
scalar called ``energy". During learning, the model learns to shape an
objective energy surface so that the modes desired in synthesized output
get assigned low energies, while the false modes are given high energies.
Supervised learning is a form of energy-based modeling in an indirect way:
for each training sample $X$ having label $Y$, the energy of the paired ($X,
Y$) is a low scalar value when $Y$ is the correctly predicted, and is a
high scalar value if incorrectly predicted. In unsupervised learning
formulation, lower energy is attributed to the data manifold represented by
input $X$. A data point which causes increase in total energy is denoted as
a contrastive sample, e.g. a false positive in supervised learning and
samples from low data density regions in unsupervised learning. The authors
propose to view the discriminator as the required energy (or contrast)
function, which has no probability-based interpretation. The total energy
computed by the discriminator is used to train the generator. Like
adversarial training, the discriminator gets trained to assign low energy
values to the modes of the data p.d.f. i.e. regions of high data density,
and higher energy values in the troughs of the p.d.f. At synthesis time,
such an energy-based generator produces samples from the regions of the
space to which the low energy is assigned by the discriminator. An
autoencoder was used as the discriminator, with the energy being the reconstruction error.
EBGAN was able to generate plausible 256$\times$256 pixel resolution images from the
ImageNet dataset.

\subsubsection{Salient Observations from Code}
\begin{enumerate}
    \item to calculate PT loss, the feature output is flattened before
taking the dot product.
    \item Discriminator is an autoencoder! Generally it is the generator which has autoencoder style architecture.
\end{enumerate}

\subsection{Towards Principled Methods for Training Generative Adversarial Networks}
\label{princ_tr_sec} A foundational theoretical investigation described in
\cite{principled_tr_pap}, into training dynamics over manifold structures
went long way into design of better GAN models. The paper specifically
studies problems including instability and saturation during GAN training.
They explain mode collapse problem using comparison of $p_r(x)$ and
$p_g(x)$. Similarly, the problem of fake-looking example has also been
explained using reverse comparison between $p_r(x)$ and $p_g(x)$. They
showed that if the supports of $P_r$ and $P_g$ are disjoint or lie in low
dimensional manifolds, one can always design a perfect discriminator, whose
gradients will be zero for all $x$. Formally, they proved that for $p_r$
and $p_g$ satisfying above conditions and being continuous on their
respective manifolds,
\begin{eqnarray*}
JSD(P_ r \parallel P_g) = \log 2 \\
KL(P_r \parallel P_g) = +\infty \\
KL(P_g \parallel P_r) = +\infty
\end{eqnarray*}

They  also similarly prove that if the other cost function of the vanilla
GAN is used to avoid vanishing gradients in the discriminator, it leads to
highly unstable updates to the discriminator weights, something that has
been widely observed practically. With these proofs, they conclude that using divergences to test and coerce distributions may turn out to be
disastrous at times.  Hence, if and when these divergences are maxed out (infinity, almost
always), attempting to minimize them by gradient descent isn't really
possible.

To fix these problems, the authors suggest to break the conditions under
which the theorems are proven. One way is to add noise to the input of the
discriminator, to smoothen the distribution of generated images/data to
some extent. As a corollary, if the two supports of the distributions $P_r$
and $P_g$ over their respective manifolds are closeby, then adding
smoothening noise terms makes the noisy distributions $P_{r+\epsilon}$ and
$P_{g+\epsilon}$ almost overlap. In such a case,  the JS divergence between $P_{r+\epsilon}$ and
$P_{g+\epsilon}$ will be small, in total contrast of the JS divergence
between $P_r$ and $P_g$. The authors prove a relation between the Wasserstein distance of
$P_r$ and $P_g$, to the JS divergence of $P_{r+\epsilon}$ and
$P_{g+\epsilon}$, and the variance of the added noise. 
$P_{r+\epsilon}$ and $P_{g+\epsilon}$ being continuous p.d.f.s, their
divergence can be be used rather to train a GAN, instead of JSD between $P_r$
and $P_g$ as the GAN loss. A discriminator trained using these
distributions will provide provably smooth and non-trivial gradients during
the training, unlike the original loss function.

\subsection{Generative Multi-adversarial Networks}
\label{gman_sec}
An earlier work before WGAN, which leads to stable training with the
original objective, uses multiple discriminators \cite{gman_pap}. For this,
they reformulate $\mathbb{G}$’s objective as $min_\mathbb{G} max_\mathbb{F}(\mathbb{V} (\mathbb{D}_1, \mathbb{G}), \ldots , \mathbb{V}
(\mathbb{D}_N, \mathbb{G}))$ for different choices of $\mathbb{F}$, where
$\mathbb{F}$ is an aggregator function and $\mathbb{V} (\mathbb{D}_i,
\mathbb{G})$ is the objective function of each vanilla discriminator. Each
$D_i$ is still expected to independently maximize its own
$\mathbb{V}(\mathbb{D}_i, \mathbb{G})$ (i.e. no cooperation). The authors
tried a number of design options, with the discriminator role ranging
from being a tough adversary to being a forgiving teacher. Compared to
standard GANs, GMAN is shown to produce higher quality samples in much
lesser time. In philosophy, it
is an ensemble approach employing multiple limited-capacity discriminators.
The authors show that averaging over these multiple locally optimal
discriminators increases the entropy of $p_{data}(x)$, by scattering the
probability mass over the vast data space.

\subsection{Unrolled Generative Adversarial Networks}
\label{unroll_sec}
Before Wasserstein GANs were introduced, a very profound method of
stabilizing GAN training was provided in \cite{unrolled_gan_pap}. Not only
did this model addressed the mode collapse problem in generated data from GANs (thus increasing mode diversity and mode coverage), but also
addressed the training instability. Unlike other works, the authors note
that mode collapse is a trade-off, with the visual sample quality of the
generated images belonging to the few modes is much better, with model
being able to focus on only the most common modes. A intuitive summary of
this work has also been provided in 
\cite{gan_tut}. As per \cite{gan_tut}, in practice, when one traces the
gradient of V ($\mathbb{G}$, $\mathbb{D}$) for both generator and
discriminator simultaneously, the max operation used in computing the gradient for
$\mathbb{G}$ gets ignored. The objective of $\mathbb{G}$ should rather be
max$_\mathbb{D}$ V($\mathbb{G}$, $\mathbb{D}$), and the back-propagation
should be through the maximization operation. Various strategies exist
to achieve that, but most of them are unstable during training, such as
implicit differentiation. Unrolled GAN attempts to solve this problem by
building a computational graph describing \textbf{k} steps of
learning in the discriminator, and then all \textbf{k} of these training
steps are used for back-propagation to calculate the generator's gradient.
Theoretically, tens of thousands of steps are needed for complete
maximization of discriminator's objective, but even 10 or fewer steps of
unrolling was empirically observed to noticeably reduce the effect of mode dropping.

\subsubsection{Salient Observations from Code}

\begin{enumerate}
    \item Equation (7) is implemented by the step loss.backward(create\_graph=True)
    \item To calculate $\mathbb{G}$'s loss, a \textbf{copy} of $\mathbb{D}$ is advanced the number of times the inner loop is unrolled.
    \item The last layer of $\mathbb{G}$ is essentially a feature map, sans any activation.
\end{enumerate}

\subsection{Geometric GAN}
\label{ggan_sec}
A generalization of McGAN was published in \cite{geometric_gan_pap}, and
considers itself as a sibling of WGAN (see subsection \ref{wgan_sec}) due
to having a common ancestor in IPM minimization. The authors observed that
McGAN's feature spaces is based on composition of three geometric operations:
\begin{itemize}
    \item Hyperplane search separation
    \item Usage of SGD to do discriminator update, away from the hyperplane
    \item Usage of SGD to do generator update, towards the hyperplane
\end{itemize}
This is a very generic interpretation, hence it can be
applied to most of the existing GAN and its variants. As per authors, the
main differences between the two algorithms are based on computation of
separating hyperplanes for a linear classifier in the embedding space, and
deriving the  geometric scaling factors for the feature vectors. Based on
this observation, the authors designed geometric GAN that uses SVM-based hyperplane
separator, which has theoretically maximal margin between two classes of separable data. The work did
not scale much.

\subsection{Unsupervised Anomaly Detection with Generative Adversarial Networks to Guide Marker Discovery}
\label{ano_gan_sec}
Anomaly detection is the task of identifying test data not fitting the
normal data distribution seen during training. The authors of a very
impactful work \cite{anogan_pap} employ unsupervised learning on
large-scale medical images' data to identify anomalous regions, that can be
construed as candidates for disease markers. Many diseases do not have a
known set of all possible markers (not all corresponding anomalous regions are available as ground truth), while in case
of some diseases, the predictive power of markers is limited i.e. the
corresponding anomaly is not profound. In the traditional anomaly detection sense, the
authors propose method to create a rich generative model of healthy/normal
local anatomical appearance. They show how GANs help in
creating a such an expressive model of normal appearance. They
propose an improved technique for creating the required latent
representation space, and use both $\mathbb{G}$ and $\mathbb{D}$ to differentiate between
observations that are similar to the normal training data, and those which
do not conform to the same.
They modify a GAN inpainting technique with two adaptations. One, they
condition the search of test image representation on the whole test image, and
two, a novel way to guide such search in the latent space is also provided (inspired
by \cite{improved_gan_pap}). An anomaly score is defined that allow robust
discrimination between normal anatomy's appearance, and locally anomalous
appearance. This approach identifies known disease markers with high
accuracy, while detecting novel markers as well for which no voxel-level
annotations are available a priori. Training is done by extracting
random patches from normal images. For testing, one is given $\langle
\mathbf{y}_n, l_n \rangle$, where $\mathbf{y}_n$ are unseen images of same
size, as of training patches, extracted from new testing data J and $l_n
\in \{0,1\}$ is an array of binary image-wise ground-truth labels, with n =
1, 2, $\cdots$, N. The learnt representation space of the generator is a
smooth manifold. Hence two points close on the manifold, when
decoded/generated, lead to two visually similar images. Given a test image
\textbf{x}, to locate anomaly i.e. a plausible marker, one needs to find an
embedding \textbf{z} in the latent space/manifold $\chi$ that generates an image
$\mathbb{G}$(\textbf{z}) that is visually most similar to test image
\textbf{x}. To find the best \textbf{z}, starting from randomly sampling
$\mathbf{z}_1$ given the fixed latent space distribution $\mathbb{Z}$, the
training process uses the trained generator to generate
$\mathbb{G}$($z_1$). Based on $\mathbb{G}$($z_1$), they define a training
loss, using which $z_1$ is updated to $z_2$. The location of \textbf{z} in
$\mathbb{Z}$ is thus iteratively optimized via $\gamma$ = 1, 2, $\ldots$, $\Gamma$
backpropagation steps. The training loss has two parts, a residual loss
and a discrimination loss. The first loss component enforces visual similarity
between the generated image $\mathbb{G}(\mathbf{z}_\gamma$) and test image
\textbf{x}. The second loss component enforces the generated image
$\mathbb{G}(\mathbf{z}_\gamma$) to have a representation that lies on the
learned manifold $\chi$. The anomaly score of the test image is then simply 
the loss value of the test image.

\subsection{Image-to-Image Translation with Conditional Adversarial Networks}
\label{pix2pix_sec}
A seminal work in image translation task using GAN was described in
\cite{pix2pix_pap}. The authors define automatic image-to-image translation
as the problem of translating one possible representation of a scene into
another, given sufficient training data. Such task always maps a high
resolution input grid to a high resolution output grid. The work known as
\textbf{Pix2Pix} uses supervised cGANs to carry out the task. Hence the
loss function is adaptive, because GANs inherently learn a loss that adapts
to the data. Unlike semantic segmentation, which considers output space to
be unstructured,  cGANs learn a structured loss and do not assume pixel
independence. The authors claim that their framework differs from prior
important works in that nothing is application-specific. For generator,
they use a “U-Net”-based architecture, and for  discriminator they use a
convolutional ``PatchGAN'' classifier, which only penalizes structure at
the scale of image patches. There are many innovations clubbed inside this
work, and can be read directly. Borrowing from an earlier, work, the
generator is tasked to not only fool the discriminator but also to be near
the ground truth output in an L1 sense, as L1 distance encourages less
blurring. The authors of \cite{bicycle_gan_pap} refer it to as ``regression
loss'' (there is also another auxiliary MTL loss in another work,
\cite{acgan_pap}, ``classification loss''). Further, instead of providing
noise at input, the authors provide noise only in the form of dropout,
applied on several layers of our generator at \textbf{both} training and
test time.  For many image translation problems, there is a great deal of
low-level information shared between the input and output, such as edge
location. Hence the authors use skip connections to shuttle this
information directly across the net, and overcome the bottleneck formation
in the general shape of a ``U-Net'' architecture. Given that L1 loss in
generator already assures correctness at low frequencies, the discriminator
can be restricted to model only high-frequency structure (see task
definition above). Hence the authors restrict themselves to patch-level
structure generation (using a MRF modeling of texture from earlier work),
which will capture much of textural variations. The real/fake discriminator
is run convolutionally across the image, averaging all responses to
provide the ultimate output of $\mathbb{D}$. It was shown that the size of
a patch can be much smaller than the image size, and still perceptually
high-quality output can be generated. The batch normalization parameters calculated
at test time are based on full batch instead of a minibatch.
\subsubsection{Salient Observations from Code}
\begin{enumerate}
    \item ResNet-34 block has been implemented in ResnetBlock class.
    \item The detailed design of both the generator and the
discriminator networks follows that described in the extended version of paper in arXiv. The
detailed architecture is in networks.py. There are some unmentioned
architectural design options also present in code (LSGAN, WGAN-GP,
ResnetGenerator), possibly used in various ablations.
    \item Visdom is used instead of tensorboard for training monitoring, so
the drawback is that there no intermediate storage of event logs into an
event file.
\end{enumerate}

\subsection{Learning from Simulated and Unsupervised Images through Adversarial Training} \label{simgan_sec}

One of the first seminal works to show the possibility of training a
classifier using simulated dataset was SimGAN \cite{mixed_learning_pap}. As
per authors, learning from synthetic images can be problematic due to
domain gap
between synthetic and real image distributions. Hence they proposed
Simulated+Unsupervised (S+U) learning, which intends to generate more
realistic synthetic images, while using unlabeled real images. Other than improving
realism, they also hypothesize that S+U learning should
preserve annotation information for training of the models. In this
approach, a synthetic image is first generated using a black box generator.
It is then refined using the refiner network. To add realism, they train a refiner
network using an adversarial loss. The refiner network must not corrupt annotations of synthetic
images, especially high-frequency information such as edges of the
annotation. Hence a self-regularization loss is used as an auxiliary loss that
penalizes any major changes between the synthetic and refined images.
To avoid introduction of unwanted artifacts while attempting to fool a single strong
discriminator, they receptive field of the discriminator was restricted to local regions only. Finally, in a similar but complementary
way to \cite{improved_gan_pap}, to avoid mode collapse, the discriminator
is updated using a history of refined images.

The self-regularization loss is a variation over pix2pix (refer section
\ref{pix2pix_sec}) that minimizes difference between a transform of the synthetic and refined images, $l_{reg}$
= $\parallel \psi(\mathbf{\tilde{x}}) - \mathbf{x}\parallel_1$, per pixel, where
$\psi$ is an image embedding, and
$\parallel$.$\parallel_1$ is the L1 norm. The refiner network i.e.
generator is a fully convolutional neural net \textbf{without} striding or
pooling operations, again to preserve annotation boundaries. For realism,
the authors hypothesize an almost-real synthetic image must have at least
patch-level similarity, just like pix2pix discriminator. Else, artefacts
may get introduced in the output in terms of few poorly matched regions (real
vs. fake) though there may be near-perfectly matched regions as well.
Working with patches in the discriminator not only limits the receptive
field, leading to less capacity discriminator network, but also
provides manyfold extra samples in form of patches, for training the
discriminator. The refiner network also behaves better and tighter, due to
patch-level `realism loss'-based training values.

\subsection{DeLiGAN : Generative Adversarial Networks for Diverse and Limited Data}
\label{deligan_sec}

Typical GAN-based approaches require large amounts of training data to
capture the diversity across the image modality. The current image datasets
used for generation have extremely complex underlying distributions. The
level of details and the diversity of the datasets leads to such
complexity. The generator network $\mathbb{G}$ needs to itself have
sufficient capacity to be able to match and efficiently learn from such
datasets.
However, a big amount of training data is required to train high-quality
generative networks. However, when data is not aplenty and yet diverse,
increasing generator's capacity is of no use. A solution
provided in \cite{deligan_pap} to this conundrum is the following: Instead
of increasing generator's capacity in form of its depth, the authors
propose to make the prior distribution of latent space sufficiently rich,
rather than a unimodal Gaussian. A reparameterization of the latent space
$\mathbb{z}$ is hence proposed, so that $p_z$, from which input to the
generator is sampled, is a multimodal Mixture-of-Gaussians \( p_z(z)~=~\sum_{i=1}^{N}\phi_i
g(z|\mu_i,\Sigma_i)\). By applying a mixture model, it is hypothesized that
one is likely to obtain samples in the other high probability regions in
the latent space as well. More importantly, the parameters of the GMM, and
that of GAN generator, are learnt together. This simple extension happens
to be quite effective strategy, and such models  are able to generate
diverse samples even when trained with limited data. It was shown to
generate images of handwritten digits, objects and hand-drawn sketches (all
diverse outputs). L2
regularizer is used in $\mathbb{G}$'s loss function to avoid learning
trivial component Gaussians.

\subsection{Plug \& Play Generative Networks: Conditional Iterative Generation of Images in Latent Space}
\label{pnpgan_sec}
Current image generative models often work well at low resolutions (e.g. 32
$\times$ 32), but struggle to generate high-resolution (e.g. 128 $\times$
128 or higher), globally coherent images for datasets with large
variability e.g. Imagenet, due to many challenges including difficulty in
training. An earlier work DGN-AM produces high-quality images at high
resolution. That model used a generator $\mathbb{G}$ to create realistic images
using latent features extracted from a pretrained classifier model
$\mathbb{E}$. To generate class-conditioned images, the model uses an
optimization process (similar to GAN inversion) to locate a hidden feature
\textbf{h} that is mapped to an image by $\mathbb{G}$ in such a way that 
it highly activates a neuron in another proxy classifier $\mathbb{C}$. The
authors point out that a major drawback of DGN-AM, however, was the
missing
diversity in the generated samples. In an important advancement to DGN-AM,
called PnPGAN \cite{pnpgan_pap}, a prior on the latent feature is added to
constrain the iterative optimization search happen on the manifold's
surface. This improves the quality and diversity of samples produced.
The constraining is done in a probabilistic framework in which the authors
interpret activation maximization objective via energy-based modeling. The
energy function they employ is a sum of multiple terms: (a)
priors such as maximum realistic look of generated images and (b)
conditions such as operating upon proxy classifier's manifold which has
been trained on at least the desired classes (and few more optionally).
The name PnPGAN arises from the advantage that the energy function can be
designed freely and differently by users, and different priors and
conditions can be ``plugged'' in and played with by the users, to come up with a
different and efficient generative model.

\subsection{Stacked Generative Adversarial Networks}
\label{stacked_gan_sec}
When the data distribution p(x) is highly variable and complex, current
generative models fail to learn it entirely. In an interesting work that deals with
this problem, \cite{stacked_gan_pap} uses a top-down stack of GANs. Each
model in the stack is trained to generate ``plausible" lower-level feature
maps that are \textbf{conditioned} on higher-level representations
generated from the previous model of the stack. To do so, a a set of
representation discriminators are introduced which when trained well,
distinguish ``fake" feature representations from ``real" feature
representations. The adversarial loss at each level, for each
representation discriminator, forces the set of intermediate
representations to lie on the manifold of a bottom-up DNN's representation
space. Each generator is trained with a conditional loss that forces it to
use the higher-level conditional input. The authors also designed a
\textbf{novel} entropy loss which forces each generator at each level to
generate diverse feature representations. The fundamental difference of
their design from a similar hierarchical GAN, \cite{lapgan_pap}, is that
both architectures comprise of multiple GANs, each working at one scale, LAPGAN
targets generation of multi-resolution images, from low to high, while SGAN
targets modeling the multi-level scale-space generative representations, from abstract to
dataset-specific. Provided with a pre-trained encoder $\mathbb{E}$ (mostly a
classifier), the goal is to train a top-down generator $\mathbb{G}$ that
inverts $\mathbb{E}$. At an intuitive level, one can decompose all the
variations in training images into multiple levels as is known in
literature \cite{cnn_vis_pap}, with higher-level variations being semantic
(e.g. shapes and category-specific parts) and and lower-level variations
being generic (e.g., edges, corners and textures). At each level, SGAN
generator uses independent noise variable/prior to represent independent
variations at that level. One loss component, conditional loss, has
been modeled on lines of few earlier works. There is one more issue: issue:
sometimes the generator $\mathbb{G}_i$ learns to ignore the noise $z_i$,
and compute $h_i$ deterministically from $h_{i+1}$. To tackle this problem,
the authors encourage the generated representation $h_i$ to be sufficiently
diverse when conditioned on $h_{i+1}$, i.e., the conditional entropy
H($h_i$|$h_{i+1}$) should be as high as possible. Since directly maximizing
H($h_i$|$h_{i+1}$) is intractable, a variational lower bound on the
H($h_i$|$h_{i+1}$) is maximized instead. Specifically, they use an
auxiliary distribution $Q_i$($z_i$|$\hat{h}_i$) to approximate the true
posterior $P_i$($z_i$|$\hat{h}_i$), and additionally use an auxiliary loss 
termed as entropy loss. The authors give a proof that minimizing
$L^{ent}_{G_i}$ is equivalent to maximizing a variational lower bound for
H($h_i$|$h_{i+1}$). This loss component is highly influenced from InfoGAN
work(see subsection \ref{infogan_sec}). There is a very similar loss
component used in a parallel work, EBGAN, as well (see subsection
\ref{ebgan_sec}). This work is not to be confused with a later work of
similar name, though more popular.

\subsection{Adversarial Variational Bayes: Unifying Variational Autoencoders and Generative Adversarial Networks}
\label{avb_sec}
While ALI \cite{ali_pap} and BiGAN \cite{bigan_pap} were indeed first to
hint towards presence of an encoder in the GAN generator, many contemporary
works including \cite{gan_vae_pap} found different ways to bring in a
encoding mechanism. The authors note the comparison of VAEs and GANs:
 while GANs generally output visually sharper synthetic images, especially
for natural images, faces etc. VAEs as a model automatically provide an
additional inference/encoding model post training, other than the generative
model/decoder. They also observed that the output of reconstruction branch
of both ALI and BiGAN only vaguely resemble the input, mainly in form of
preserving semantic identity and not pixel-level intensities. The posterior
distribution learnt by the encoder within VAEs is apparently not expressive
enough, to be sampled and decoded into sharp images, a drawback of VAEs. To
overcome this, the authors present a VAE training mechanism having
arbitrarily flexible encoders parameterized by neural networks. They used
AVB to train a generator to sample from a unnormalized probability density.
In contrast to a standard VAE with
Gaussian latent space, they include the noise $\epsilon_1$ as additional
input to the encoder instead of adding it at its output, which allows the
VAE encoder to learn complex probability distributions.

\subsection{Conditional Image Synthesis with Auxiliary Classifier GANs}
\label{acgan_sec}
In a seminal work, authors of \cite{acgan_pap} demonstrate via empirical
analysis that high resolution samples embed class-specific information that
is not present in low resolution samples. They also showed that a more
structured GAN latent space and better losses lead to generation of higher
quality images. As per their analysis, a high number of classes to be
synthesized is what makes ImageNet synthesis difficult for GANs. To solve
this problem, they hypothesize that instead of providing any class-specific
conditional information to the discriminator, the discriminator should
rather be tasked with reconstructing such class-specific conditional
information. They do so by extending the discriminator to have an
auxiliary classification head that outputs the class label for the training
data, or related latent variables presumably encoding the class identity.
The overall framework employs multi-task learning naturally \cite{mtl_pap}.
The additional classification head can optionally use a separate
pre-trained encoder of its own (e.g. classification backbones), which
further improves the quality of generated output. Though ACGAN is also
class-conditional generator by definition, it also has the auxiliary
classification branch. The could show that their high-resolution output was
not classical upsampling/interpolation, but much richer output indeed. In a
first, they demonstrated synthesis for all 1000 ImageNet classes at a
128x128 spatial resolution, which was higher at that time. Though a shade
less performance than its contemporary PnPGAN(see subsection
\ref{pnpgan_sec}), this work has been much more popular, most likely due to
simplicity of its design.

\subsubsection{Salient Observations from Code}
\begin{enumerate}
    \item Generator uses class conditional information to have a loss head called auxiliary generator loss, as per equations (2) and (3) of paper.
    \item Design of targets/ground truths to calculate two discriminator loss heads is quite involved.
    \item Usage of CUDNN for GPU-optimized fast filter operations.
\end{enumerate}

\subsection{Learning to Discover Cross-Domain Relations with Generative Adversarial Networks}
\label{discogan_sec}

A popular work just preceding the seminal cycleGAN work, with same
intentions, i.e. to make image translation task unsupervised, appeared in
\cite{discogan_pap}.  Hence it improves upon the earlier supervised
version, pix2pix (see subsection \ref{pix2pix_sec}). DiscoGAN addresses the
problem of estimating cross-domain relations when presented with unpaired
data in two domains. The relations are estimated with an objective to
preserve semantic attributes such as orientation and identity, while
transferring the style from one domain to another. Like cycleGAN, two
different GANs are coupled together. Each of
them ensures that corresponding generator maps, in forward and reverse
direction respectively, one domain to the other. The authors impose a
bijective condition i.e. all images in one domain are assumed to be
transferred from all images from the other domain. Using bijectivity and
invertibility, domain translations are well covered in both directions.

There seems to be \textbf{simply no difference} between the formulations of
DiscoGAN and cycleGAN, other than that DiscoGAN has been tested on simpler
datasets, and its source code is publicly unavailable. A note in
\cite{unsup_trans_pap} says that these two works discovered exactly the
same thing independently. Unlike cycleGAN which uses L1 loss, DiscoGAN
claims any of the L1, L2 or Huber loss can be used. The encoder uses leaky
ReLU, while the decoder uses normal ReLU.

While the invertible mapping is designed to remove the mode collapse
problem, the authors provide a beautiful example about how invertible
mapping from domain \textbf{A} to domain \textbf{B} alone does not solve
the problem. They then show that adding another invertible mapping, from
domain \textbf{B} to domain \textbf{A} leads to a proper bijection, and
hence reduces the mode collapse to a great degree.

\subsubsection{Salient Observations from Code}
\begin{enumerate}
    \item Code contains implementation of feature matching loss: image\_translation.py
    \item generator and discriminator parameters are chained together, so that they could be optimized simultaneously
\end{enumerate}

\subsection{Wasserstein GAN}
\label{wgan_sec}
Probability density models that are tractably fit using MLE is not
efficient in commonplace datasets where the data distribution is defined
over low-dimensional manifolds. This is because the support of the model
manifold and that of the data manifold are unlikely to have a non-negligible
intersection. In such a case, the KL divergence term is undefined or
infinite. In both VAE and GAN frameworks, instead of estimating a parametric
$\mathbb{P}_r$ which is unlikely to exist, a random variable \textbf{Z}
with a parametric distribution $p(z)$ can be used as input and fed to a
parametric generative function estimator $g_{\theta}~:~\mathcal{Z}
\rightarrow \mathcal{X}$, which directly outputs samples from the data
distribution $P_\theta$. $\theta$ can be learnt iteratively so that
$P_\theta$ converges towards the real data p.d.f. $P_r$. In a seminal work, the
authors of \cite{wasserstein_gan_pap} first comprehensively and
theoretically analyze, how the Earth Mover (EM) distance compares against various well-known probability
distances and divergences generally employed in generative modeling.
They prove that the Wasserstein distance is much weaker than the JS distance used
in GAN framework.  Using that, a new model called Wasserstein GAN is
introduces that minimizes an approximation to EM distance, using
Wasserstein distance. They theoretically prove that usage of WS distance is
optimal when it comes to coercing one p.d.f. ($P_\theta$) to another ($P_r$).
It was further shown experimentally that WGANs cure the main training problems
of GANs. In WGAN, one need not carefully design and balance the capacities
of the generator and the discriminator networks. Also the choice of
architectural units within each network e.g. the normalization unit etc. is
not very critical in Wasserstein GAN. WGAN attempts to ameliorate the mode
dropping problem in GANs from first principles, and hence the drops are
drastically reduced. Unlike vanilla GAN and derivates, where practically
the discriminator cannot be trained till optimality but only till fixed
number of epochs, in WGAN, one can do so. The output samples are also of
much higher quality.  To implement EM distance,
\textbf{Lipshitz constraint} has to be imposed on the network weights, which was
approximated using weight clipping in this paper.

\subsubsection{Salient Observations from Code}
\begin{enumerate}
    \item Leaky ReLU has been used with BN. Batch normalization itself is
optional, and a hyperparameter is provided to turn it off.
    \item At test time, output of G which is representing a fake image, has pixels in the range [-1,1]. They are renormalized to range [0,1], and then saved as image.
    \item Most GANs have ratio of \textbf{iterations} per epoch for
alternating training of $\mathbb{D}$ and $\mathbb{G}$. Hence one of them
consumes one minibatch, while the other consumes multiple minibatches as
per the implementation.
    \item Even with WGAN, when parameters of $\mathbb{G}$ are being
updated, subgraph representing $\mathbb{D}$ is frozen, as is the case with
all GANs.
\end{enumerate}
`   
\subsection{DualGAN: Unsupervised Dual Learning for Image-to-Image Translation}
\label{dual_sec}
Like DiscoGAN \cite{discogan_pap}, another contemporary work to cycleGAN
having similar approach is \cite{dualgan_pap}. The authors make a
\textbf{startling claim}: depending on the task complexity, thousands to
millions of labeled image pairs are needed to train a conditional GAN.
Instead, the authors use \textit{dual learning} paradigm to translate
images, which is lighter on data requirement. In dual learning, the
training entails two agents that play a dual-learning game, each of
whom only understands one image domain+one language domain, and can find out the
likelihood of the translated images/sentences being inlier
images/well-formed sentences in targeted image/language domain. For learning/backpropagation, two objectives are used:
the membership score which is the likelihood of the translated samples
belonging to the target domain, and the reconstruction error that
measures the distance between the original and reconstructed sample. Both
objectives are realized using the domain details. Surprisingly, they use
asymmetric model in form of \textbf{FCNs} as domain translators. The
discriminators have  Markovian PatchGAN architecture, like pix2pix. The authors
of the paper are honest about the superiority of results of cycleGAN.
However, their work is better than Coupled GAN work(refer subsection
\ref{coupled_gan_sec}), since the weight-sharing assumption in CoGAN does
not always hold. Hence CoGAN's applicability is not general but
task-dependent, leading to unconvincing image translation results, as shown in
comparative studies by CycleGAN. For training the model, they use Wasserstein
distance, since it has better performance both in terms of output quality and
convergence of the generator, and avoiding mode collapse.
They use RMSprop instead of Adam, the most common solver for GANs.

\subsection{Unpaired Image-to-Image Translation using Cycle-Consistent Adversarial Networks}
\label{cyclegan_sec}
Unsupervised Image-to-image translation task pertains with understanding
the individual properties of a set of images, and learning to translate
these properties into another set of properties characterizing a different set of
images. This has to be done the absence of any paired training examples,
across the sets. This can lead to e.g.
converting an image from one visualization of a given scene, x, to
another, y, while preserving the scene identity/semantics. Examples of such
translation include edge map to full image, labels to image (semantic image
synthesis), grayscale to color (colorization) etc. To avoid requirements of
large datasets for training, the authors of a seminal work
\cite{cycle_gan_pap} assume that some inherent mapping between the two
domains, and then make the model learn that mapping. They
exploit supervision at the level of sets. In other words, the trained model learns
a mapping $\mathbb{G} : X \rightarrow Y$, such that the distribution of
images from $\mathbb{G}(X)$ is indistinguishable from the distribution $Y$
using an adversarial loss. Because this mapping is highly
under-constrained, one needs to reduce the space of possible mapping
solutions. So they bias the model selection and couple it with an inverse
mapping $\mathbb{F} : Y \rightarrow X$, and bring in a novel cycle consistency
loss to coerce $\mathbb{F}(\mathbb{G}(X)) \approx X$ (and vice versa). The
counter-cycle consistency loss is also applied, i.e.
$\mathbb{G}(\mathbb{F}(Y)) \approx Y$, to ensure bijection. The authors do
not assume that the input and output domains have to map to the same
low-dimensional manifold. In cycleGAN, they use two adversarial
discriminators $\mathbb{D}_X$ and $\mathbb{D}_Y$, where $\mathbb{D}_X$
targets distinguishing between images of one domain {$x$} and corresponding translated images
{$\mathbb{F}(y)$}; in the same way, $\mathbb{D}_Y$ targets discriminating
between images from the other domain{$y$} and corresponding translated
images {$\mathbb{G}(x)$}. The use least-squares objective and
SimGAN \cite{mixed_learning_pap} strategy of gradient update using batch
history.

\subsection{Least Squares Generative Adversarial Networks}
\label{lsgan_sec}
Authors of a very important work \cite{lsgan_pap} found that the binary
cross entropy loss, used in the discriminator of traditional GAN,
can result in the vanishing gradients problem during the model training.
Specifically, they show that usage of BCE loss leads to vanishing gradients
for the generator parameters, especially with respect to true positive fake
samples, but are not close to the real-vs-fake decision boundary. In GAN,
however the objective is not have large margins, but the opposite: we want
good fakes to be as close to real samples and hence the far-away fakes must
be pulled close to the decision boundary, during the game. To solve this
problem, they take a simple approach to employ and additional least squares
loss. This additional loss is able to move the faraway true positive fakes
towards the decision boundary, for the simple reason that such loss
penalizes fake samples that have large distances/disparities with real
samples i.e. lie far away from the decision boundary. Hence this loss not
only retains proper classification of fakes, but also narrows down the
decision margin as desired. Hence such a GAN is able to generate more
realistic samples i.e. better
quality images. Further, while fakes with robust margin lead to miniscule
gradient, fakes with robust margin lead to larger least squares distance
and hence when they are put together, the overall gradient value for the
generator is much higher. This ameliorates the vanishing gradient problem
in vanilla GAN. A contemporary work, WGAN (refer subsection
\ref{wgan_sec}), as a secondary contribution, evaluated GAN training
stability by dropping batch normalization from the architecture. The
authors of LSGAN corroborated that LSGANs are also able to converge to a
relatively good state \textbf{without} needing batch normalization. They follow the
DCGAN architecture exactly.

\subsection{ALICE: Towards Understanding Adversarial Learning for Joint Distribution Matching}
\label{alice_sec}
A work by the same authors as that of ALI (see subsection \ref{ali_sec}),
provably improves the image quality performance over the seminal work,
cycleGAN. Possibly the latter are still highly popular over ALI's sequel,
ALICE \cite{alice_pap}, since it is easier to train. It was shown that
ALICE had much more training stability than ALI, and the output is much
more diverse and realistic. They show that there is an 
identifiability problem with the solutions to the ALI objective. In order
to reduce the impact of the problem, they put constraints on the
conditional probabilities denoted as $q_\phi(z|x)$ and $p_\theta(z|x)$. The authors introduce conditional
entropy as an auxiliary loss and prove that the cycle consistency, another
auxiliary loss in cycleGAN, is an \textbf{upper bound} of the conditional
entropy loss. So ALICE formulation is tighter than cycleGAN. In general,
cycle-consistency, in DiscoGAN and cycleGAN, $L_{cycle}(\theta,\phi)$, is
implemented via $l_k$ losses, for k = 1, 2, and real-valued data such as
images. As a consequence of an $l_2$-based pixel-wise loss, the generated
images generally turn out to be blurred. After identifying this problem, to overcome the
reconstruction problem, the authors use cycle-consistency using
\textbf{fully} adversarial training, instead of $L_{cycle}(\theta,\phi)$. Specifically, to
reconstruct $x$, they specify an $\eta$-parameterized discriminator
$f_\eta(x,\hat{x})$ to distinguish between $x$ and its reconstruction
$\hat{x}$. Equivalence of ALI and cycleGAN was also looked at theoretically,
and it was demonstrated that ALICE further improves the quality of output
while using adversarially learned cycle-consistency.

\subsection{Bayesian GAN} 
\label{baygan_sec}
A lesser-known work does hyperparameter search over generator's parameters,
a standard way of Bayesian inference in deep networks
\cite{bayesian_gan_pap}. To avoid mode collapse, various works change
various divergences behind the two-player game. The authors of this work
point out that it is tough to decide the type of divergence to used in GAN
training for various image domains. As per them, one should rather try
doing a fully probabilistic inference. This leads to Bayesian modeling in
which a posterior distribution over the generator parameters is assumed,
which can be multimodal and broad. The two-player game based traditional
GAN training learns this whole posterior distribution as a point mass
located on a single mode, not all modes. This leads to standard mode
collapse problem, even if the training setting is not about model
overfitting. Assumption of a multimodal posterior also entails the
interpretation of multiple different virtual generators, each leading to
different type of synthesis. Assuming a fixed full representation for the
posterior over both generator and discriminator parameters, one can model
and match the data distribution more accurately. One can also perform
data-efficient semi-supervised learning on top of it. The drawback of this work is that
it works with smaller/easier datasets such as celebA, MNIST, CIFAR, and is
also targeted to semi-supervised learning.

\subsection{Toward Multimodal Image-to-Image Translation} 
\label{bicyc_gan_sec}
In an important work, authors of \cite{bicycle_gan_pap} note that many
image-to-image translation problems are ambiguous, as a single input image
may correspond to multiple possible outputs. At the same time, training
datasets for I2I task provide bijectively mapped pairs only. When naively
used, it was shown by them that pix2pix and other similar models generate
high-quality images at the cost of diversity/multi-modality. This happens
because the generator surprisingly learns to ignore much of the elements of
the random noise vector at its input, when conditioned on a relevant
context/input image $A$. They quote some papers that show that ignoring the
noise leads to more stable training, but diversity is compromised. To overcome
this, the authors model the output of translation as a distribution over
plausible outputs, in a conditional generative modeling setting.
Specifically, in BicycleGAN, they learn this true conditional distribution
given input image $A$ from which they sample an output $\hat{B}$ at
synthesis time. Such sampling generates images that are both diverse and realistic. This is the first
work which talks about \textbf{night-to-day} image translation problem.
This work also imperatively improves upon pix2pix \cite{pix2pix_pap}, though they do refer its
sequel, cycleGAN, as well.  Their generator contains both an encoder and
a decoder(strictly speaking, encoder is an attachment to the decoder i.e.
generator which decodes the latent vector into a fake image). They
integrate two of the earlier approaches, one including InfoGAN
\cite{infogan_pap}. To learn the mapping, they learn a low-dimensional
latent space $z \in \mathbb{R}^Z$, which models the \textit{probabilistic} aspects
of the output sample which are not present in the input image. For example,
given a sketch of a handbag, the synthesized output could be in a variety
of plausible colors and textures, and this knowledge gets represented in
the learnt  latent code. They then learn a deterministic mapping
$\mathbb{G} : (A, z) \rightarrow B$ to the output. To enable stochastic
sampling in an easy way, the latent $z$ thus learnt should ideally be drawn
from some prior known distribution $p(z)$ such as a Uniform or a Gaussian 
distribution (they used Gaussian distribution).

\subsection{Fisher GAN} 
\label{fishgan_sec}
Another work which attempts to improve the training stability of GANs is
Fisher GAN \cite{fisher_gan_pap}. They claim to be an \textit{improvement} on
WGAN-GP (see subsection \ref{wgan_gp_sec}), a relatively much used
framework. Like Geometric GAN, they build upon an earlier work, McGAN.
In WGAN-GP, a penalty is imposed on the gradients of the critic. In Fisher
GAN, a constraint is imposed rather, on the second-order moments of the
critic's output. They
introduce Fisher IPM, a scaling invariant distance between distributions.
They prove certain important properties of this distance. Fisher IPM is
then as a GAN objective. Their training algorithm that uses this objective
shows desirable properties: a stable, lightweight loss between distributions
for GAN as in Wasserstein GAN, , that is easy to implement. The capacity of
critic is not compromised, only the objective is simplified. In
semi-supervised generative settings, Fisher GAN achieves strong results,
and as some of the earlier works, \textbf{avoids} usage of batch
normalization in the critic.

\subsection{GANs Trained by a Two Time-Scale Update Rule Converge to a Local Nash Equilibrium}
\label{ttur_sec}
In a very important work towards proving the \textbf{convergence} of GAN
training, the authors of \cite{ttur_gan_pap} propose a two time-scale
update rule (TTUR) for training GANs with \textit{stochastic gradient
descent} on arbitrary GAN loss functions. As per this rule, they motivate
usage of independent learning rates for the generator and the
discriminator. They point out the flaw in the recent theoretical
understanding of GAN convergence: the expectation values are computed over
training samples going to infinity, which is impractical since any model
training involves mini-batch learning, and also stochastic gradients.
Models that are trained using two learning rate rule provably converge
under easy assumptions, to a local and stationary Nash equilibrium point. In fact, the
convergence also happens if Adam optimization is used instead of SGD, which
(Adam) is less popular but more robust. The authors also introduce a new
metric to measure synthesis quality, ``Frechet Inception Distance''(FID).
It measures the similarity of synthesized images to real images, and is
shown to be \textit{better than} the Inception Score on that. TTUR, when
used with DCGANs and WGAN-GP, outperformed conventional GAN, when trained
over many popular datasets. The intuition behind TTUR is that when the
generator (weights) is fixed, the discriminator trains and converges to a local minimum.
However, if the generator weights change very slowly, then the
discriminator network still provably converges, since the changes in generator output will be small deltas.
While doing so, the discriminator is forced to first learn new patterns,
before they are used to train the generator. On the contrary, if the
generator learning is made too fast, it will force the discriminator
steadily into new regions, before it can learn about the previous region. In recent GAN implementations, e.g. \cite{wasserstein_gan_pap,
improved_gan_pap}, the generator indeed learnt much slower than the discriminator.

\subsection{Unsupervised Image-to-Image Translation Networks}
\label{unsup_trans_sec}
While cycleGAN improved upon CoGAN among many other GAN frameworks on image
translation task, an enhancement of CoGAN, called UNIT
\cite{unsup_trans_pap}, claims to subsume and supersede cycleGAN's
performance. Since the image correspondence is not provided, the Unsupervised
Image-to-image Translation (UNIT) problem is a hard problem. However, it is
more generic and hence practically useful, since collecting images without
establishing correspondence is much easier.
In such problem formulation, there are two sets consisting of images
belonging to two marginal distributions defined over two different image
domains. The translation task is then to infer the joint distribution using
these training images, which can be used to do translation of new images
later. A theory called \textit{coupling theory} in probability states that
an infinite set of joint distributions exist which can lead to a given
marginal distribution. Hence, given a marginal distribution, establishing a
fixed joint distribution is a highly ill-posed problem. As always, to
address this ill-posed problem, one needs to constrain the solutions using
additional assumptions on the form of the joint distribution. The authors
make a shared-latent space hypothesis as the constraint, which implies that
images that have a correspondence due to an optimal translation network in
between, share a common representation space, and have same representation
within it. The weight-sharing constraint in latent space is imposed along with
adversarial training objective, to generate corresponding images in two
domains, while the VAEs map translated images with input images in their respective domains. They show that the
\uline{the cycle-consistency constraint is implied by the shared-latent
space constraint}, but the vice-versa is not true. Hence the latter is a stronger hypothesis. Hence it can be reasoned that shared-latent
space assumption is a less generic assumption than cycle consistency
assumption. However, shared latent space assumption impacts the
architecture design, while cycle consistency constraint impacts the loss
function design. Hence the authors used both, and show by ablation, that
usage of more constraints did lead to better quality results, as against
using either one of the two. This work also talks about night-to-day
conversion dataset.

\subsection{Triple Generative Adversarial Nets}
\label{tripgan_sec}
There are two main problems in existing GANs for SSL: (1) the generator and
the discriminator play game but do not converge to their optimal point
together; and (2) the generator does not impact the semantics of the fake
outputs, since real training samples are only available to the
discriminator. Works such as \cite{improved_gan_pap} and
\cite{ttur_gan_pap} try to work on the first problem, but do not work well
in semi-supervised setup. In \cite{improved_gan_pap}, the feature matching
loss improves works well in classification but can lead to mode collapse,
while the minibatch discrimination loss produces realistic outputs, but is
not satisfactory at predicting labels accurately. As per authors of
\cite{triple_gan_pap}, such performance tradeoff happens since the (single)
discriminator has to perform two incompatible tasks -- predicting class
labels and discriminating fake images. The authors also showed that none of the existing GANs actually learn the
disentangled representations in SSL, other than limited few works, including ACGAN \cite{acgan_pap}.
To handle these problems in a semi-supervised setup, the authors introduce
two new conditional networks -- a generator network to generate pseudo
images from real labels, and a classifier to do pseudo labeling for the
real images. To synthesize realistic output while using these conditional
networks, they employ a single discriminator network which only does one
task: it determines, in a supervised way, if a synthesized data-label pair is different from the
any pair within real labeled dataset or not (inlier determination).
It is named Triple-GAN because not only are there are three networks in
action, but also that there are three joint distributions in the
background: the true data-label distribution and the distributions
defined by the conditional networks. In particular, they prove that instead
of trading off in general adversarial networks as stated in the first
problem, a good classifier in such setup will result in a good generator
and vice versa. Also, to solve the second problem, the discriminator can use the (pseudo-)label
information of the unlabeled data as generated by the classifier, to make
the generator synthesize properly paired images and labels.

\subsection{Improved Training of Wasserstein GANs}
\label{wgan_gp_sec}
The authors of \cite{wasserstein_gan_pap} themselves note that 
WGAN generates poor fake images sometimes, or fails to converge.  WGAN
requires that the discriminator (also known as critic) must be a
k-Lipshitz function, which is implemented indirectly via weight clipping.
It was found out that this strategy leads to optimization difficulties.
Specifically, even with batch normalization usage in the critic, it was
observed that very deep critics fail to converge many a times during
training. Thus this strategy entails careful tuning of the clipping
threshold as a hyperparameter, $c$. Even if WGAN converges, this strategy
has a side effect in the sense that it biases the critic towards
discriminating much simpler generator functions. In
\cite{wpgan_pap}, the authors propose a different method to implement
Lipshitz constraint. The authors directly limit the gradient norm of the
critic’s output, with respect to critic's input. To make this limiting operation
tractable, a soft version of the constraint is also provided, that
penalizes the gradient norm for random samples $\hat{x} \sim P_{\hat{x}}$,
against the limit.
They argue and recommend layer normalization as a drop-in replacement for
batch normalization, at least in the critic. The also argue in favor of
two-sided penalty, unlike e.g. \cite{improved_gan_pap}.

\subsection{A Note on the Inception Score}
\label{inc_score_sec}
An important evaluation \cite{inception_score_note} brings out various
shortcomings of the earlier GAN evaluation metric, IS. According to them,
not only the metric is suboptimal, but also has issues with its
application. High-dimensional deep generative models, unlike e.g. KDE, are
very tough to evaluate. Meaningful metrics is needed to evaluate and track
the model improvements. Inception store is an ad-hoc evaluation metrics
used towards such a goal. Since the latent representations signify the
empirical generative data distribution being learnt, a way to measure
output quality could be to use a pretrained backbone network over the generated images, and then
calculate certain statistics in the representation space.
Inception Score, Mode Score and Frechet Inception Distance take this
approach. An ideal score belongs to a model that can generate varied and
realistic images, and has high visual quality. However, if the generative
model overfits and memorizes the (large) training data, these metrics fail.
For example, they show that the Inception Score is sensitive
to minor perturbations in network weights, though any classification accuracy of
such metric does not vary much. A good metric for evaluating generative
models is not expected to be sensitive to changes, when the generated
output itself does not change much. One way to achieve this while using
Inception Score is that the Inception Network that is used in this scoring
must be trained using the dataset that was used by the generative model. Thus the
original Inception Score must be used for measuring natural image generators.

\subsection{Generative Adversarial Networks: An Overview}
\label{gan_overview_sec}
A fantastic \textbf{starting point} to understand GANs is the well-written survey paper, \cite{gan_overview_pap}. It covers most of the important work till end of 2017, by which time GAN training had reached certain maturity.

\subsection{Data Augmentation Generative Adversarial Networks}
\label{dagan_sec}
In many realistic settings, one needs to train models with limited
datasets. In those cases, deep neural networks run into known problems,
tending to overfit on the training set, leading to poor generalization. One
known way to reduce overfitting is to use data augmentation. As per authors of
\cite{dagan_pap}, standard traditional augmentation methods provide only
limited alternative data that is feasible. Specifically, typical data augmentation
techniques use a very limited set of known (affine) invariances that are
easy to invoke. The authors argue that a bigger space of invariances can be
learnt, by training a form of conditional GANs in a different data-rich
domain, known as the source domain. This space can then be used for
augmentation in the target domain which entails low amount of training data. A good property is that DAGAN does not depend on the
classes themselves. The performance advantage of DAGAN holds even when
augmentation space is transferred between \textit{substantially separate} source and target domains. The crux of the
work is in the philosophy that a GAN learns transformations mapped on a
data manifold: $z$ = $\theta$ gives one point on the manifold, and
perturbing $z$ in each different direction along the tangent plane
continuously maps out to latent representations of augmented samples on the
data manifold. On the architecture front, the work curiously uses batch
\textit{renormalization} and DenseNet for discriminator architecture.

\subsection{Spectral Normalization for Generative Adversarial Networks}
\label{specnorm_sec}
Another \textit{foundational} work towards stabilization of GAN training
was provided in \cite{spectral_norm_pap}. It introduces a novel weight
normalization technique called spectral normalization to stabilize the
training of the \textbf{discriminator}. Spectral normalization is easy to
implement and lightweight, in the sense that one needs to only tune the Lipshitz constant hyper-parameter. The authors showed that 
the discriminator training amounts to the finding an unbiased estimator for the
\textit{density ratio} between the data distribution and model
distribution. In high dimensional spaces formed by e.g. image
representations, this estimation is often unstable and inaccurate. This
leads to the well-known failure of generator to learn the multimodal
structure of training data, i.e. mode collapse.  The authors argue towards
the effectiveness of spectral normalization for GANs against other Lipshitz regularization techniques, and thus \textit{improves upon} WGAN and WGAN-GP. 

\subsection{cGans with Projection Discriminator}
\label{pdisc_sec}
Most existing discriminator architectures in cGANs take in the conditional
signal $y$ into the discriminator by naively concatenating
$y$ to the input signal $x$, or to some feature map of an intermediate
layer.  In \cite{cgan_proj_pap}, the authors studied the conditional
probabilistic models as represented by the discriminator. Briefly, a
BCE-trained discriminator approximates a function that minimizes the information
theoretic distance between the p.d.f.s of synthetic data and target data. Any hypothesis about the form of the target/data distribution
acts as a regularization, that impacts the discriminator design. Looking
this way, the authors propose a specific discriminator form, which is motivated by a probabilistic model in which the
conditional distribution $y$ given $x$ is either discrete or uni-modal
continuous distribution. Such assumption commonly lies behind many
real-world applications, such as superresolution and class-conditional image
synthesis. Enforcing this hypothesis leads to a structure of the
discriminator that entails an inner product of the latent condition vector $y$ and
the feature map. On superresolution task, they were able to produce high
quality super-resolved images, which showed better classification accuracy
downstream, than the cGANs that use non-learning based bicubic and
bilinear interpolation methods. In class conditional
image generation, this model produced significantly better images when
trained on ILSVRC2012 (ImageNet) 1000-class image dataset, as compared to
state-of-the-art methods, without using multiple generators or
discriminators i.e. limited capacity. Some examples of simple 
$p(y|x)$ and $q(y|x)$ distributions that they use are are latent log linear
and latent Gaussian distributions. However, they also
point out that if the situation calls for multimodal $p(y|x)$, it might be
sensible to avoid using this model. The work argues to have improved upon ACGAN
(see subsection \ref{acgan_sec}), and PnPGAN(see subsection
\ref{pnpgan_sec}).

\subsection{Progressive Growing of GANs for Improved Quality, Stability, and Variation}
\label{prog_gan_sec}
In classical GAN formulation, when one measures the distance between the
training distribution and the generated distribution, the gradients can
point to more or less random directions if the distributions do not have
substantial overlap, i.e., are too easy to tell apart. As per authors of
\cite{acgan_pap}, thigh-resolution images are tough to generate
because at that resolution, it is easy to discriminate the generated images
from real images at train time, a severe form of above gradient problem.
Also the intermediate tensors consume lot of GPU memory and hence only
smaller minibatches can be used at training time, which affects stability
of training. A way to address this problem, in a very influential paper
\cite{progressive_gan_pap} is that one should progressively increase the
sampling interval/grow discriminator and generator functions so that they
output low-resolution images initially. Newer layers can then be added to
both, and the model retrained, so that higher-resolution details get
modeled via these new layers that are added iteratively and trained. In ProgressiveGAN, the authors use
generator and discriminator architectures that mirror each other and grow
together harmoniously. Due to incremental training, the model first
discovers the coarse, global approximation of the target distribution, and then slowly
learns fine-grained details of the distribution at various scales, instead of 
needing to learn all details at all scales simultaneously. By increasing
the output resolution slowly in multiple steps, essentially the training
objective of learning the complex generative p.d.f. is partitioned into a
series of simpler objectives. Simpler objective not only increase the
trained speed and convergence, but also improve training stability in high
resolutions. Mode collapse can still happen in such model, and it starts
when the discriminator learns faster, leading to high gradients, and hence
needs to be carefully controlled. The loss function employed by
ProgressiveGAN is quite flexible in choice, from LSGAN-loss \cite{lsgan_pap} to WGAN-GP loss
\cite{wpgan_pap}. As such, the work betters two contemporary works
(sections \ref{hres_sec} and \cite{madgan_pap}).

\subsection{AttnGAN: Fine-Grained Text to Image Generation with Attentional Generative Adversarial Networks}
\label{attn_gan_sec}
Most recently proposed text-to-image synthesis methods encode the whole
text description into a global sentence vector as the condition for
GAN-based image generation. However, authors of \cite{attngan_pap} note
that such global conditioning omits important word-level fine-grained
information, which degrades the quality of generated images. For
scene-centric images having multiple salient objects, this problem in
generation is even more severe, e.g. while training with the COCO
dataset. To address this issue, in \cite{attngan_pap}, the authors propose
an multi-stage attention-based refinement for detailed text-to-image
synthesis. The model consists of two novel components. The first component
is an attention-based generator, in which a novel attention block within
the generator synthesizes different sub-regions of the image by
shortlisting and interpreting words that most likely relate to the sub-region being drawn.
Specifically, other than employing a global sentence vector derived from
the scene description, each word is further embedded into a word vector.
The generator first generates a low-resolution image using the global
vector. In the subsequent stages, it uses each image sub-region's
representation along with an attention layer to query word vectors, leading
to formation of a word-context vector. The region-specific image
representation is then combined with the word context vector, leading to a
multimodal context vector. This is used by the model to generate new image
features in the neighboring sub-regions. Such multi-stage strategy results
in a higher resolution image getting synthesized at each stage.

\subsection{High-Resolution Image Synthesis and Semantic Manipulation with Conditional GANs}
\label{hres_sec}
Another important work that targets synthesis of large images
(2048$\times$1024) is \cite{img_synth_pap}. The process of graphics
engine-based rendering is turned into a model training and inference problem. The
authors build on \cite{gman_pap} and \cite{cascade_gan_pap}. New multi-scale
discriminator and multi-scale low-resolution-to-high-resolution generator submodels are
proposed, that are amenable to very high resolution conditional image generation. They
use a novel adversarial loss as well. Furthermore, they extend the framework to
interactive visual manipulation with two additional features. First, they
allow object instance segmentation information as conditional information,
to be able to do object manipulations viz. change object category and
remove/add objects. Second, another method is provided to ensure diversity
in synthesized output, so that users can manipulate the object appearance interactively.

\subsection{DeblurGAN: Blind Motion Deblurring Using Conditional Adversarial Networks}
\label{dbgan1_sec}
Inspired from usage of GANs in many low-level vision tasks such as
super-resolution and inpainting, the authors of \cite{deblurgan_v1_pap} use
GAN for motion deblurring as well. Deblurring is modeled as a special form
of image-to-image translation using GANs. They use WGAN-GP (see subsection
\ref{wgan_gp_sec}) additionally with perceptual loss. WGAN-GP is used since
its advantage is retained irrespective of the generator architecture, and
the authors of DeblurGAN wanted a lightweight generator for a preprocessing
step. They train on a \textbf{mix} of real and simulated blurred images.

\subsection{Generate To Adapt: Aligning Domains using Generative Adversarial Networks}
\label{dom_al_sec}
A problem in varying amount of labeling is pointed out in \cite{acgan_pap},
wherein the authors tell that across different data domains, while labeled
data is available, much more than preceding years, the amount of labeling
and its distribution differs for each data domain/dataset. Due to this
difference, even the best state-of-the-art models have inferior performance on
real-but-unseen test-time data. If one however uses the unlabeled data from
target domain to adapt the model (domain adaptation), the shift between source and target can
be mitigated. In \cite{gan_dom_align_pap}, a representation that is robust
to the domain shift is learnt by a model. It is based on shared feature
space assumption. Such embedding is directly learnt jointly from the source
domain labeled samples and target domain unlabeled samples. The work builds upon ACGAN
\cite{acgan_pap}. During training, the source domain images are fed to an
encoder, and the representation thus produces is used not only by a
classifier for supervised label prediction, but also by a generator to
synthesize a realistic source domain image. During backpropagation, the
updates happen both from the discriminative gradients from the
classifier and generative gradients from the adversarially trained
generator. In the adaptation stage, the encoder is updated using only gradients from
the adversarial generator, since the target domain data is devoid of
labels. Such adapted encoder is hypothesized to retain its discriminative
classification capability even in the target domain, due to its earlier 
generative-discriminative training. The discriminator being 
a multi-class classifier, the discriminator gradients for the batches of
unlabeled target images guide the encoder representations towards feature
subspace of the respective classes. By \textit{sampling from
the generator distribution during synthesis}, they show that the
network has indeed learned to bring the source and target distributions
closer. Due to multi-task learning, unlike the previous methods, their
domain adaptation process can also train the model when in cases
where image generation part is quite hard, especially the low-data image
generation scenario. While for low-data datasets, synthesized new samples
generally only exhibit style transfer, for domain adaptation, it is shown
to be enough for providing good gradient information, that successfully
aligns source and target domains using their approach.

\subsection{From source to target and back: Symmetric Bi-Directional Adaptive GAN}
\label{sbada_sec}
In unsupervised domain adaptation problem, there is rich amount of
annotated source domain data on which to train a deep
network, but the target domain annotations are not available. One
option is translating the source domain samples into samples from the target domain, either by
changing the image embeddings, or by directly synthesizing a new type of
source images. A similar but inverse model translates the target domain
samples into the source domain. As per authors of SBADAGAN \cite{sbada_gan_pap},
given a fixed generative model, the direction of domain adaptation is
application-driven: there may be cases where translating from the
source domain to the target domain is easier, and cases where it is true otherwise.
SBADAGAN uses two adversarial losses that encourage the model to output
target-looking images from the source images and
source-looking images from the target images. Moreover, it jointly minimizes
two classification losses. Further, it uses the source domain classifier to
produce the label of the source-like translated target images. Next, it forces a
new semantic condition on the source images using the class consistency loss.
It imposes that by translating source images towards the target domain and then
again back towards the source domain, which forces that the samples should become very close
to the class of their ground truth. This condition is more relaxed
than a typical reconstruction loss, as it uses only image annotations'
data, and not any appearance information. Still, their experiments show that it is highly
effective in aligning the representations of the source and targets
domains. The work hypothesizes that the source domain and target domain consist of the same set of
classes, and it uses the process of self-labeling.

\subsection{StarGAN: Unified Generative Adversarial Networks for Multi-Domain Image-to-Image Translation}
\label{stargan_sec}
Existing approaches towards image-to-image translation have limited
scalability and robustness in handling more than two domains, since
different models should be built independently for every pair of image
domains. In certain cases, different domains are correlated in the sense
that they differ in certain attributes while the underlying object class is
the same. This often happens in case of face datasets where collection of
face images based facial expressions(`happy', `angry', `sad' etc.), hair color, age, gender etc. can be deemed
as a domain in itself. In such cases, attribute can be defined as a
meaningful feature representing an object within an image, such as color of
apple (red/green), and can have discrete or continuous values. For such
correlated domains, novel tasks such as multi-domain image-to-image
translation can be defined, where one can change image from one domain
according to attributes sourced from multiple other domains.

If domain translation models were to be learnt among these many domains,
then one would naively need to train $k(k-1)$ generators. Such generators
cannot interoperate with each other since even if they learn pairwise
shared, the features aren't truly global and do not impact other domains.
This renders the usage of these models for the new task ineffective, and
the quality of generation is also not up to the mark. Further, only one
attribute per domain is labeled, so the label space for each domain is also
partial only.  StarGAN \cite{stargan_pap} learns joint mapping among all available
partially-labeled domains using only one generator. The generator takes an
additional input about the domain encoding, rather than assuming a fixed
target domain. Thus it learns to flexibly translate an input image into
one of the multiple different domains, as desired at runtime. The desired
domain is encoded as a onehot vector. During training, the target domain
label encoding is randomly generated, so that translation to all domains
can be learnt in a flexible way. The method can also handle
domains that span across different datasets. Finally, it is able to deal
with partial labeling problem by using a mask of one-hot vector encoding of
label information. Instance normalization is used within the generator but no
normalization is used within the discriminator.

\subsection{Augmented CycleGAN: Learning Many-to-Many Mappings from Unpaired Data}
\label{aug_cycle_sec}
As per \cite{aug_cyclegan_pap}, a weakpoint of the seminal cycleGAN
\cite{cycle_gan_pap} model is that it assumes that the mapping between
source and target domains is approximately deterministic and bijective.
Especially when source and target domains are distant, e.g. faces with
expressions and faces portraying certain age groups, and cycleGAN does not
produce good results in such cases.  Hence mapping across
domains must not be assumed to be that simple, and a better assumption would
be to relax and assume many-to-many mapping. To implement this hypothesis, 
the authors add additional latent variables to each domain's
representation, and use extended training procedure of cycleGAN to these
extended latent spaces. The input to their model is not just a source
domain image, but also a latent variable, and the output is another latent
variable along with a target domain image. More specifically, a mapping
between pairs $(a,z_b) \in A \times Z_b$ and $(b,z_a) \in B \times Z_a$ is learnt, where $Z_a$ and $Z_b$ are
extended latent spaces that model any missing information when translating
from domain $A$ to domain $B$, and vice-versa. 
Due to use of additional
variables, the mapping might be one-to-one in the augmented space after
training, but after marginalization of the additional latent variables, it
becomes many-to-many.  They use two cycle-consistency
constraints to model the loss. They also use conditional instance
normalization to incorporate multiple style generation, for style
transfer.

\subsection{CyCADA: Cycle-Consistent Adversarial Domain Adaptation}
\label{cycada_sec}
The problem of \textbf{unsupervised domain adaptation} implies that one is
provided source data $X_S$, source labels $Y_S$, and target data $X_T$ ,
but no target labels. The goal is to learn a model $f_T$ that correctly
predicts the label for the target data $X_T$. \textit{Feature-level}
unsupervised domain adaptation models perform domain adaptation by aligning
the feature space of source domain as extracted by a network, with that of
target domain. The source and target domains could be synthetic and real
image domains, respectively. Such aligning requires a loss minimization of
some form of distance between the two feature spaces and their
distributions. Such adaptation faces two main problems. First, the aligned
distributions are marginal distributions, and that is oblivious to any 
semantic consistency, e.g. feature space of birds may get mapped to feature
space of elephants. Second, alignment done at e.g. higher representation
levels leave the aspects of low-level appearance variance unmodeled, and
that can degrade the model performance. If one switches to generative
\textit{pixel-level} domain adaptation models, they undertake similar
alignment of distributions. But such image-space models are limited to
alignment of small-sized images and limited domain distances. Also, cycleGAN was not designed for
domain adaptation problem. An important work that tries to take middle path
on this problem is CyCADA  \cite{cycada_pap}. This model coerces representations at
both the feature-level and pixel-level, and also enforces semantic
consistency. Specifically, both structural and semantic consistency is enforced during domain adaptation using a cycle-consistency loss, and
task-specific semantics losses. The semantics losses ensures discriminative
power of the overall representation, as well as semantic consistency of the
output w.r.t. target domain.

\subsection{How good is my GAN?}
\label{howgood_sec}
An ImageNet-pretrained Inception network is used in both IS and FID
measurement of performance of generative models, especially GAN. This is
not appropriate for generative models trained on datasets from other domains
such as medical images and faces, which are not similar to Imagenet at all.
While both usefully measure relative advancement and training stabilization
of the generative models, the final performance of the model on unrelated
real-world datasets is not properly measured by them. As an alternative,
one can calculate the distance of the generated images to the real images, 
and further, generative precision and recall \cite{gan_study_pap}. For
generative models, high precision implies that the generated samples are
mostly inliers i.e. near the real data manifold,
and high recall implies that  the generated samples cover most of the data
manifold. Manifold of natural images is complex and mostly unknown, so
these two derived measures are not computable in theory.
Indeed, the evaluation in \cite{gan_study_pap} is limited to using
synthetic data composed of gray-scale triangles. A different measure called 
 sliced Wasserstein distance (SWD) was introduced in
\cite{progressive_gan_pap}, to evaluate GANs in proper way. It is an approximation of 1-Wasserstein
distance between generated and real images, and given the Laplacian pyramid
nature of Progressive GAN, is computed using statistical similarity between
local image patches extracted at various scales. The work in \cite{howgood_gan_pap}
on \textbf{evaluation measures} betters SWD. They propose new evaluation
metrics to compare class-conditional GAN models with separate train and
test scores. To compute GAN-train score, they train a classification network
using GAN generated images, and then evaluate its classification performance on a test
set having only real-world images. Intuitively, this metric approximates the
difference between the generative/learnt and the real/target distributions.
GANs are prone to mode collapse, so a robust GAN-train value implies good mode
coverage from real distribution (less modes are missed), and in that sense,
can be construed as some form of a recall measure. To compute GAN-test
score, they train a classification network \textit{inversely} using real
images, and then evaluate its performance on a test set made of only 
generated images. A robust GAN-test value implies that the generated images
are inliers i.e. realistic enough and almost sampled from the real image
distribution (i.e. mostly true positives), and in that sense, can be
construed as some form of a precision measure.
They also study the utility of generated images for augmenting training
data for downstream models. Experiments with these two metrics demonstrate
that the \textit{quality of GAN generated images decreases significantly as
the complexity of the target data distribution increases}.

\subsection{Are GANs Created Equal? A Large-Scale Study}
\label{gan_eq_sec}
The authors of \cite{gan_study_pap} conduct a third party, large-scale
multi-objective empirical evaluation of state-of-the art GAN models using
the available evaluation measures. As per them, the main problem with
direct evaluation is that one cannot \textit{explicitly} compute the
generative probability $p_g(x)$, due to usage of neural network consisting
of non-linearities. Due to this, classic generative measures, such as
log-likelihood on a test set, are inapplicable. Hence most research works
focus on qualitative evaluations involving the visual quality of samples,
using experts. Such measures are subjective and can exhibit observer bias,
and be misleading as well. The authors empirically demonstrated that almost
all popular GANs attained similar values of FID measurement, given long
training runs and larger minibatch sizes. GAN models with lesser
hyperparameters were found to have performance advantage over GAN models
with more hyperparameters. Evaluation measure-wise, Inception score was
shown to mainly captures precision while, FID measure was shown to capture
both precision and recall. In a different way, the authors defined a series
of generative tasks, whose complexity increased in an almost-smooth and controlled
way, and used the generated samples to indirectly measure precision and recall (which are well-defined bias-free metrics). Thus they
were able to compare different models based on \textit{well-known} metrics.
Given the target data manifold which signifies the distribution $p_d$, the
authors employed a distance measure such that on efficient computation of
the distance of each generated sample from $p_d$, one can aggregate and
compute some sort of precision and recall for effective performance measurement of the
generative model. Towards measuring precision, if the generated samples
from distribution $p_g$ are closer to the manifold, then the precision is
deemed higher (also see section \ref{howgood_sec}). Similarly, if the
generator could generate samples that cover almost all of the manifold
$p_d$, then the recall is higher.  However, as subsection \ref{howgood_sec}
notices, the manifold is created using toy datasets, a major limitation of
this comparison.

\subsection{GAN Dissection: Visualizing and Understanding Generative Adversarial Networks}
\label{gan_dis_sec}
The authors of \cite{gan_vis_pap} design a framework to understand GAN
outputs hierarchically at the part-, object-, and scene-level. In this
framework, they attempt some questions. Which internal tensors in GAN model
give rise to synthesis of perceivable objects? Are these tensors just
correlated to the object synthesized, or can generate the object in
isolation/independently? How is context in the scene represented with GAN
model? In the framework, the authors first identify a group of
interpretable architectural units having some relation to
object concepts. These tensors corresponding to these units correlates well
with the segmented region of a particular object class (e.g. doors).
Another set of architectural units is identified empirically whose presence
or absence can cause an object to get synthesized or not. This effect is
measured using a standard causality metric. Finally, the relation between
the background and the set of these causal object units is examined. The
location of insertion of synthesized object regions and any shift in
location of other objects in the overall synthesized scene is studied. This
framework is then used to compare layer-wise representations in GANs, trained
on different datasets. Given a layer's tensor \textbf{r}, the framework examines whether
\textbf{r} explicitly represents any class $c$ in some way. One can also use it to locate and remove spurious
artefacts that get synthesized.

\subsection{Large Scale Gan Training for High Fidelity Natural Image Synthesis}
\label{biggan_sec}
In a path-breaking effort, authors of BigGAN \cite{biggan_pap} attempt the
problem of successfully generating high-resolution, diverse samples from
complex datasets such as ImageNet, which has been elusive. For GANs, the
standard problem at any resolution is that of the gap between realism of generated
images (mode replication), and the variety (mode coverage), and the authors
attempt to bridge this gap. The authors show
that scaling the network size benefits the synthesis tremendously. Hence
they employ models having upto four times as many trainable parameters, and
upto eight times the minibatch size, as compared to earlier models.
They use SAGAN (see subsection \ref{sagan_sec}) as the baseline model. They
provide class information to $\mathbb{G}$ with class-conditional BatchNorm
and to $\mathbb{D}$ with projection \cite{cgan_proj_pap}. They showed that
just increasing the minibatch size itself gives a lot of boost to
generative model's performance. As per them, this most likely happens since 
each bigger batch covers more modes at training time, and hence better
mode-covering gradients to both generator and discriminator. Further,
increasing the network width i.e. the number of channels in each layer
leads to further performance improvement in the generative model (twice the
number of parameters leads to around 20\% increase in inception score).
This is attributed to fitting: increased model capacity w.r.t. dataset
complexity improves mode replication. Improving upon LapGAN, 
direct skip connections from the input noise vector $z$ to multiple layers
of $\mathbb{G}$ are also added and tried out, with an intuition to allow the generator $\mathbb{G}$ to
use the latent space to \textit{directly} influence generator's features at
different resolutions in the multi-resolution scale space. This was done by grouping elements
of $z$ into one group per scale, and concatenating each group to the
conditional vector $c$ at each layer. In another orthogonal improvement, taking a model trained with z $\sim \mathcal{N}(0, I)$ and sampling $z$ from a truncated normal distribution
immediately improved generation performance. Controlling the truncation
allowed them to balance the earlier mentioned tradeoff. To counteract saturation artefacts
introduces by usage of truncated normal, the authors use various forms of
orthogonal regularization. Their model can generate up to 512$\times$512
resolution images, and in a much simpler fashion than that in ProgressiveGAN
\cite{progressive_gan_pap}.

\subsection{The Relativistic Discriminator: A Key Element Missing from Standard GAN}
\label{relgan_sec}

While the initial successful GAN models were based on f-divergences, recent
models such as Wasserstein GANs are based on Integral
probability metrics (IPMs). The discriminator of such GANs is real-valued
and constrained to approximate a specific class of functions so that it
does not learn fast. In a sense, it ($D$) is regularized so that its
gradients do not die out too fast. Spectral normalization as an IPM
constraint can be used to constrain f-divergence based GANs as well.
However, the authors of \cite{relativistic_gan_pap} argue that
non-IPM-based GANs actually lack something else that IPM-based GANs have: a relativistic
discriminator. This component is in fact shown to be necessary to make GANs
perform minimization of divergence, and do sensible classification knowing
that half of the samples in the current minibatch to the discriminator are
in fact fake samples. Such modified GANs are more stable during training,
and produce more realistic samples. During the two-party game, this
discriminator naively tries to classify most samples as real, after generator's
training, which is not right since only half samples are fake. During
divergence minimization, training should not only increase D($x_f$) but also decrease D($x_r$).
However, if D($x_r$) always decreases when D($x_f$) increases,
discriminator loss' gradient will get influence by the real data. Such
modified GANs were able to train
in very difficult settings, in which traditional GANs get stuck in training early
itself. 

The ``relativistic discriminator" actually estimates the posterior
probability that the randomly sampled fake data is less realistic than 
given real data, sample-wise or on average. Both of these give rise to 
non-standard GAN loss functions, and the GANs that use it are called and refer to them
Relativistic GANs (RGANs) and Relativistic average GANs (RaGANs)
respectively. IPM-based GANs are later shown to be a subset of RGANs.
RaGANs especially are able to synthesize 256x256 high-resolution realistic
images, given low-shot training data (upto 2011 training samples), which
the f-divergence based GANs such as LSGAN cannot. Further, their realism
was found to be significantly better than the ones generated by WGAN-GP \cite{wpgan_pap} and
SNGAN \cite{spectral_norm_pap}.

\subsection{Self-Attention Generative Adversarial Networks}
\label{sagan_sec}
The authors of an important work \cite{self_attngan_pap} found that
convolutional GANs can model some classes easily, but face difficulty which
modeling the mode corresponding to other classes. Their hypothesis for this
observation is that such models rely heavily on convolution operator to
model the contextual dependencies in the image. Due to small-sized, local
receptive field, the long-range contextual information gets processed only
in the last few, deeper layers, in case a deep generative model is used.
Self-attention is known to have capability of modeling long-range
dependencies, at least in discriminative models. It computes the attention map at a location as a weighted sum of
the features at \textbf{all locations}. The weights, known as attention
weights, can be computed in a cost-efficient way. Their work, SAGAN, introduces a
self-attention mechanism into convolutional GANs. The self-attention module
acts in a \textit{complementary} way to convolutions within the generator,
and models multi-scale long-distance contextual dependencies between image
regions. Such a attention-augmented generator synthesizes images in which
the fine-grained details Armed with at each position are well-synchronized
with fine-grained details in distant regions of the image. Given long-range
modeling, even the discriminator can employ complex geometric conditions
on the global structure on images, using self-attention. SAGAN works seems to be seminal, given the way the
results have been quoted, and even BigGAN sees to be based on it now.
However, possibly it is task-specific, i.e. meant for image synthesis task.
They use spectral normalization \cite{spectral_norm_pap} as well as TTUR
\cite{ttur_gan_pap} techniques.

\subsection{A Style-Based Generator Architecture for Generative Adversarial Networks}
\label{stgan_sec}

Authors of \cite{style_gen_pap} came up with a novel generator
architecture in which one can control the image generation. The generator
inputs a learnt \textit{constant}, and controls the ``style" of the
generated image at each layer using the latent code. This controls the
strength of scale-specific features of a generated image. Further, for
stochasticity, noise is also added into the lower layers of the network, at
each pixel level independently, so
high-level attributes of the generated image such as identity, pose get separated from low-level
stochastic variations such as freckles in an unsupervised way. This also
leads to mixing of scale-wise features and interpolations. The input latent
vector is added to latent space at intermediate levels by the generator,
which impacts which factors of variations of generated images get
represented. The latent representation of input exhibits some degree of
entanglement of attributes, that cannot be avoided. However, the
generator's intermediate latent representations are controlled
independently and hence are in general disentangled in terms of style
attributes. Overall, the factor of
variations of the generated images have very less entanglement and are
almost linearly represented in the output. Being a style-based generator,
any change in style at any layer manifests into changes in the entire
generated image since entire feature maps are deconvolved into generated
images, not regions of it. Especially the global attributes such as
background texture, shade and pose can be easily controlled in this model.
The pixel-level noise that is added does not impact synthesis of global
attributes, which is penalized by the discriminator if such thing happens.
Thus style synthesis and control does not need any supervision as such.

\subsection{AutoGAN: Neural Architecture Search for Generative Adversarial Networks}
\label{autogan_sec}
Most earlier GANs have an architecture wherein generators and
discriminators are shallow networks. Shallowness was needed due to the
training instability problems of GAN. Recent generator networks employ deep
residual networks, such as SAGAN, SNGAN, WGAN-GP, BigGAN, which have not
just training stability but also generate 
high-resolution images. The work \cite{autogan_pap} employs NAS on SNGAN to
finalize the architecture of the generator and the discriminator within.
The work does not improve the IS or FID of contemporary schemes.

\subsection{DeblurGAN-v2: Deblurring (Orders-of-Magnitude) Faster and Better}
\label{dbgan2_sec}
Improving on earlier work, \cite{deblurgan_v2_pap} is much faster in
execution while maintaining performance, raising hope of real-time
deblurring. In discriminator, they use relativistic discriminator
\cite{relativistic_gan_pap} with least-squares loss. In generator, they use
FPN architecture, and use patches.

\subsection{Seeing What a GAN Cannot Generate}
\label{gan_gen_sec}
While GANs are prone to mode collapse/drop, not much research has tried to
find out which modes get dropped by GANs during training.
In \cite{gan_seeing_pap}, the authors focus on identifying such modes by
looking at generated output. They try to find out those difficult images
that are ignored by GANs during training, due to which model distribution
does not faithfully mimic the target distribution. They also try to find
out if some specific regions of a training image (e.g. object part or
specific objects within a scene) are ignored during training. Generally
generative models train with face, object or scene datasets. For scene
images, the authors analyze the distribution and instances to locate mode
drop. For that, both real images and generated images are segmented and
compared. This way, one can detect which object or which part of object was
dropped. Then one can find the closest real image to a generated but
deficient image, and compare at intermediate layer representation level to
establish a cause for dropping of specific classes.

\subsection{Improved Precision and Recall Metric for Assessing Generative Models}
\label{prec_rec_sec}
When modeling a complex manifold for sampling purposes, two separate goals
emerge: individual samples drawn from the model should be faithful to the
examples (they should be of “high quality”), and their variation should
match that observed in the training set. The popular metrics e.g. FID and
IS do not retain this tradeoff but merge it into a single value. The
authors of \cite{prec_rec_gan_pap} present a new performance
metric that separately measures both the aspects of mode coverage and mode
replication in generative models reliably, using separate
\textbf{non-parametric} manifold representations of the synthetic and real
images. The notion of precision and recall in image generation
was introduced couple of years back. In generative modeling, precision
indirectly measures quality of mode replication while recall indirectly
measures degree of mode coverage of the target distribution. The authors
also bring out
the weaknesses of these concepts, and design and present improved precision
and recall metrics. Then then evaluate the new metrics on StyleGAN and
BigGAN.

\section{Conclusion}
In this survey, we have tried to cover the chronological aspect of research
progress in generative adversarial network modeling. Generative adversarial
networks still remain popular in many applications, where the training data
is aplenty. The advent of diffusion modeling as another form of explicit
generative models has taken away some focus from GANs. However, the theory
of diffusion modeling is still evolving and many issues including slow,
iterative inference remain a bottleneck. In contrast, the theory of
GAN modeling is now mature, though few biases and limitations of it are
expected to remain forever. The chronological development of this research
topic in generative modeling also provides a sketch of priority order in
which problems in future generative models can be tackled. In fact, some of
the problems in diffusion modeling are being tackled in almost similar
order. It is in this way that this survey tries to bring value to the field
of generative modeling.

\bibliography{ref}

\end{document}